\documentclass{article}
\PassOptionsToPackage{table}{xcolor} 

\usepackage{booktabs} 
\usepackage[preprint]{corl_2026} 

\usepackage{setup}

\begin{document}
\title{Structured Perception Representations for Off-road Dynamics Learning}

%


\author{
  Micah Nye\textsuperscript{1}, Samuel Triest\textsuperscript{1}, Matthew Sivaprakasam\textsuperscript{1}, \\
  \textbf{Isaiah Adu\textsuperscript{1,2}, Sebastian Scherer\textsuperscript{1}, Wenshan Wang\textsuperscript{1}} \\
  \textsuperscript{1}The Robotics Institute, Carnegie Mellon University, \textsuperscript{2}Pennsylvania State University\
}
\maketitle

\begin{center}
\includegraphics[width=\linewidth]{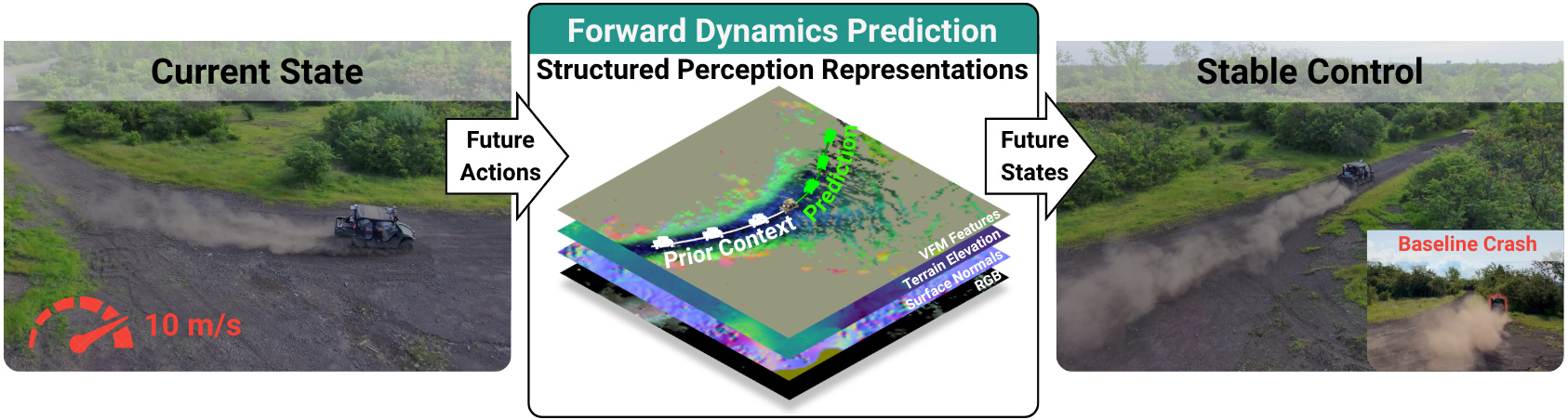}
\captionof{figure}{A learned dynamics model, grounded by structured perception representations. By leveraging the spatial structure of bird's-eye-view maps, we are able to decrease worst-case trajectory prediction error by \worstCaseErrDecrease\% from learned and classical baselines at speeds in excess of 10 m/s.}

\label{fig:title_fig}
\end{center}

\begin{abstract}
     Autonomously navigating off-road terrain at high speed demands accurate control, which often relies on dynamics models that capture complex interactions between the robot and terrain. In off-road regimes, terrain interaction is a major component of the vehicle's dynamics. Prior work attempts to capture this effect by training learned models of the vehicle's dynamics, conditioned on its perception of the local environment. However, said conditioning is often overly simplistic, ignoring the spatial structure of the dynamics problem. Our approach, SpatialBEV, exploits this spatial structure. At each step, we query the bird's-eye-view (BEV) features around the predicted state, conditioning the next prediction on current geometric and semantic cues. We additionally present a benchmarking approach that mines a dataset's most difficult scenarios. On datasets collected across seasons, SpatialBEV reduces prediction error by \worstCaseErrDecrease\%, compared to physics-based and learned baselines. We further evaluate on hardware on high-speed ($\geq$10m/s) and geometrically challenging courses with a 25.3\% reduction in maximum cross-track error. More details can be found at https://theairlab.org/spatial-bev-dynamics.
\end{abstract}

\keywords{Learning for Control, Dynamics Modeling, Perception Representations, Off-road Navigation} 

\section{Introduction}
Autonomous navigation in unstructured, off-road environments at high speeds remains a significant challenge for model-predictive control (MPC) frameworks. MPC is a well-established method in control theory; however, its success depends on accurate dynamics models for forward prediction. Traditional methods rely on simplified physics models \cite{han_model_2024, seegmiller_vehicle_2013, howard_optimal_2007, gibson_multi-step_2023, kelly_vector_2014} that struggle to capture complex terradynamics induced by challenging terrains like loose gravel, deep snow, or standing water. High speeds exacerbate these unmodeled interactions which may lead to task failure.

While learned dynamics models have emerged as a potential solution due to their expressivity, they often neglect environment observation \cite{ostafew_learning-based_2014, chua_deep_2018}. When a robot commands identical actions under identical initial conditions on different surfaces with differing geometry or friction, the resulting trajectory varies significantly \cite{dupont_effect_1993, bode2007learning}. In state-only models, this variance is captured as an unmodeled disturbance term or stochastic noise term. However, this uncertainty is largely epistemic; it is a direct consequence of the model's inability to observe the environmental physics imposed on the robot during interaction.

Motivated by this shortcoming, we propose a dynamics representation that conditions its prediction on environment geometry and visual semantics. We use birds-eye-view (BEV) representations which provide a spatially-aligned structure for environment attributes such as terrain elevation and Visual Foundation Model (VFM) features. While prior works encode BEV features as a fixed contextual prior at the beginning of a prediction pass \cite{roth_learned_2025, lee_learning_2023, lupu_magic-vfm_2024}, this leaves the spatial representation underutilized. Instead of treating the BEV information as a single prior, we make full use of its spatial structure by treating it as a continuous source for incremental feedback and state correction at every prediction step. This explicit architecture ensures the most relevant BEV features are queried at each step. This is especially important at high speed, where a robot traverses large distances and fixed-prior approaches condition every prediction step on the same initial encoding, rather than the features local to the robot's current position.

We demonstrate the efficacy of this representation through extensive evaluation. We also introduce a systematic testing methodology that clusters similar actions to isolate high-variance regimes, specifically targeting the scenarios where current models are likely to fail and where environment context is necessary for accurate dynamics prediction. Our results span both offline evaluation on aggressive driving data around multiple test sites as well as online hardware deployment on high speed ($\geq$10m/s) and geometrically challenging courses. Through all experiments, we show that our spatial representation for incremental conditioning reduces forward prediction error and improves closed-loop tracking. Specifically our work provides three key contributions:
\begin{enumerate}
    \item A new method for leveraging structured perception representations for environment context in dynamics modeling
    \item A benchmarking approach to mine the most challenging sequences of dynamics datasets
    \item Extensive offline experiments and online hardware experiments at high speeds ($\geq$10\,m/s), demonstrating the efficacy of our approach
\end{enumerate}


\section{Related Works}
Both the learning objective and representations for the dynamics model are critical for robust control on varying terrains. \citet{djeumou_autonomous_2023} choose to learn the parameters of a tire model, which grants a structured learning paradigm with natural saturation in physically infeasible regimes, but also hinges on the accuracy of this physical assumption. Such models struggle on off-road terrains, with difficult-to-model nonlinearities imposed by undulated terrain surfaces, which \citet{gibson_dynamics_2024} address by learning the force residual to a tire model. Dynamics residual learning supplements the physics model by learning any unmodeled disturbances and is used widely in literature for off-road platforms \cite{gibson_multi-step_2023, lupu_magic-vfm_2024, lee_learning_2023, levy_meta-learning_2025, ostafew_learning-based_2014, maheshwari_piaug_2023}, on-road platforms \cite{xue_learning_2024, djeumou_one_2024}, and aerial vehicles \cite{mohajerin_deep_2018, oconnell_neural-fly_2022}, among other embodiments. Yet, these approaches depend on the physics model accuracy, require ample time to tune the physics model, and restrict the model to an explicit physical form factor.

Thus, much like model-based reinforcement learning \cite{wang_rough_2022, kahn_badgr_2020, nagabandi_neural_2018}, it is common to learn the forward states directly, without any physics-informed prior \cite{triest_tartandrive_2022, kumar_rma_2021, ward_online_2025,ingebrand_zero-shot_2024, atreya_high-speed_2022, radosavovic_humanoid_2024, hwangbo_learning_2019, tremblay_multimodal_2021}. These state representations can be further formulated probabilistically \cite{fan_deep_2020, chua_deep_2018, zhang_integrating_2018, koppel_online_2016, seegmiller_vehicle_2013, compton_dynamic_2024, levy_learning_2024, doerr_probabilistic_2018}. To improve terrain awareness, dynamics models can be augmented with geometric and semantic inputs using perception representations. Prior work conditions dynamics models on first-person-view (FPV) images \cite{kahn_badgr_2020, bar_navigation_2025, hafner_dream_2020, chen_identifying_2024}, point clouds \cite{kim_learning_2022}, or BEV representations, specifically elevation maps \cite{roth_learned_2025, gibson_dynamics_2024, lee_learning_2023}, RGB maps \cite{pokhrel_cahsor_2024, maheshwari_piaug_2023}, or VFM feature maps \cite{lupu_magic-vfm_2024, gibson_dynamics_2024}.
However, most prior work treats perception as a fixed context, encoded into the latent state once at rollout start \cite{roth_learned_2025, lupu_magic-vfm_2024}. \citet{lee_learning_2023} reduce the encoder load via a BEV crop around a conservative feasible motion extent. Most similar to our approach, \citet{gibson_dynamics_2024} query features at tire points along the rollout; however, they use this context as a physics residual and no further. In contrast, our approach takes full advantage of the context by using it for both state prediction \textit{and correction}, which contributes substantially to prediction accuracy. Beyond the architecture, we formalize this spatially-grounded representation and isolate its benefit through extensive ablations and quantitative \& qualitative evaluations.

\begin{figure}[t]
    \centering
    \includegraphics[width=1.0\linewidth]{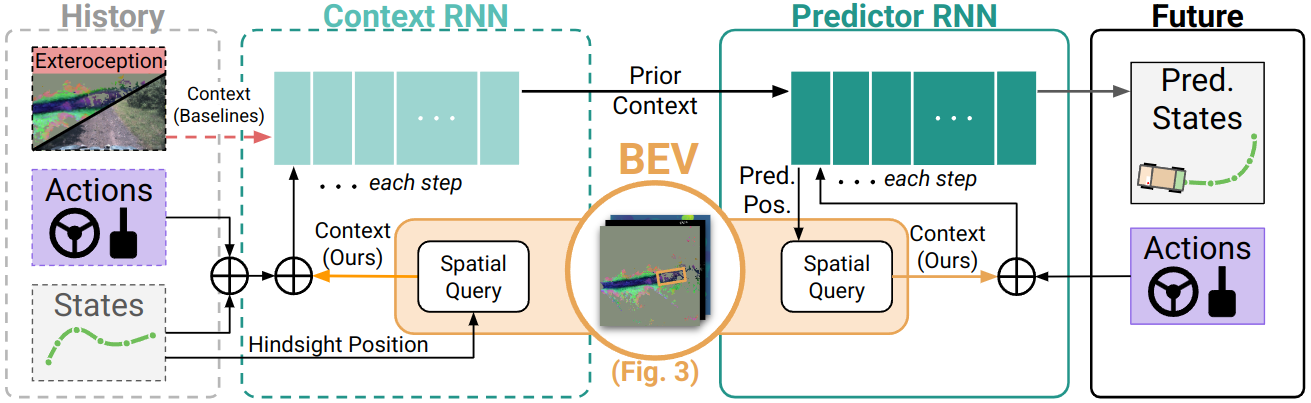}
    \caption{An overview of our dynamics model. We query each prior or predicted position in the static, structured BEV map at each step. This provides geometric and semantic context to our model for more accurate dynamics rollouts. \Cref{fig:rec-bev-cell} details the structured representation formally.}
    \label{fig:arch}
\end{figure}

\section{Method}
We aim to formulate a dynamics model that leverages perception to more accurately predict forward motion over challenging terrain before traversing it. We first detail the proprioceptive forward dynamics backbone (Sec.~\ref{sec:method-fdm}), then describe our method for leveraging BEV representations that better exploit spatial structure (Sec.~\ref{sec:method-percep}). Figures \ref{fig:arch} and \ref{fig:rec-bev-cell} illustrate our approach.

\paragraph{Preliminaries}
Given a current state $\textbf{\state}_t \in \mathcal{\State}$ and action $\textbf{\action}_t \in \mathcal{\Action}$ the next state $\textbf{\state}_{t+1}$ is governed by the deterministic robot dynamics $f_\theta$ such that $\textbf{\state}_{t+1} = f_{\theta}(\textbf{\state}_t,\textbf{\action}_t)$. For a sequence of $T$ actions forward in time $\textbf{\action}_{t:t+T}$, $f_\theta$ predicts the future states $\textbf{\state}_{t+1:t+T+1}$. We augment the model with proprioceptive context of $\history$ prior timesteps of state and action history~$\context_t=[\textbf{\state}_{t-\history:t},\textbf{\action}_{t-\history:t-1}]$. Additionally, to capture terrain and environment conditions not encoded in state alone, the robot receives a local observation $\observation_t\in\mathcal{\Observation}$ from onboard sensors, such as a BEV map. Together, the robot dynamics follow $\textbf{\state}_{t+1} = f_{\theta}(\textbf{\state}_t,\textbf{\action}_t, \context_t, \observation_t)$.

\subsection{Proprioceptive Dynamics Model}
\label{sec:method-fdm}

We architect the core dynamics model as a Recurrent Neural Network~\cite{rumelhart_rnn_1987}. Similar to \citet{gibson_multi-step_2023}, we compose it of two components: 1) the predictor network composed of RNN cells $\predrnn$ which takes an action sequence with prior context to predict the future states and 2) a context network composed of RNN cells $\contextrnn$ which initializes the prior context for the predictor network through state history, action history, and current observations. We choose to use an RNN rather than a Transformer \cite{vaswani2017transformer} because its sequential recurrence lets us insert our structured spatial representation between cells, while keeping the memory footprint small.

The predictor cell for time $t$ takes action $\textbf{\action}_t$, encoded as $\textbf{\rnninput}_t$, as input with prior latent state $\textbf{\rnnstate}_t$ to predict the next hidden state and recurrently produces this output for $\textbf{\action}_{t:t+T}$~(eq.~\ref{eq:forward-pred}). Like \citet{nagabandi_neural_2018}, we avoid predicting state derivatives for numerical integration. Instead, we predict the state differences $\Delta \textbf{\state}_t$ over a time step, so the dynamics function follows~
\smash{$\textbf{\state}_{t+1} = \textbf{\state}_t+f_\theta(\textbf{\state}_t,\textbf{\action}_t,\context_t,\observation_t)$} (eq. \ref{eq:forward-pred}).
This localizes the target near zero for a better learnable signal.

We provide proprioceptive context to the predictor through the context network cells $\contextrnn$. Taking the state and action history $\textbf{\state}_{t-\history:t}$ and $\textbf{\action}_{t-\history:t-1}$, the context network initializes the predictor's hidden state $\textbf{\rnnstate}_0$ (eq. \ref{eq:context-init}) and, importantly, trains the shared state and action encoders. While purely proprioceptive architectures like this suffice in structured domains with benign terrain, off-road environments with undulating terrain and variable ground cover need perception for greater information on the terrain ahead.
\begin{equation}
\label{eq:forward-pred}
    \textbf{\rnninput}_t=\encaction\left(\textbf{\action}_t\right),
    \quad
    \textbf{\rnnstate}_{t+1}=\predrnn\left(\textbf{\rnnstate}_t,\textbf{\rnninput}_t,\context_t\right),
    \quad
    \text{for }t=0,\dots,T,
    \quad
    \textbf{\state}_{t+1}=\textbf{\state}_t+\encstate\left(\textbf{\rnnstate}_{t+1}\right)
\end{equation}
\begin{equation}
\label{eq:context-init}
    \textbf{\rnnstate}_{k+1}=\contextrnn(\textbf{\rnnstate}_{k},\textbf{\rnninput}_{k},\context_{k}),
    \quad
    \text{for } k=-\history,\dots,-1,
    \quad
    \textbf{\rnnstate}_{-\history}\triangleq\vec{0}
\end{equation}

\subsection{Structured Perception Representations for Environment Context}
\label{sec:method-percep}
Perception remains an underutilized modality for dynamics prediction, despite its ability to expose otherwise unobserved semantic and geometric priors that affect future robot locomotion \cite{bode2007learning}. Prior work \cite{roth_learned_2025, maheshwari_piaug_2023, lee_learning_2023, lupu_magic-vfm_2024} has incorporated perceptual conditioning through inputs such as FPV images or BEV maps as fixed context signals. These approaches do not properly exploit the spatial structure of the BEV representation; we can leverage the spatial organization to get more informative, localized context at each prediction step. Through structured BEV representations, we embed geometric cues in the form of terrain height estimation and normal layers, alongside visual semantic features from VFMs. We construct these representations through a streamlined perception pipeline relying on LiDAR pointclouds for geometric features and map registration and FPV imagery for semantic features, detailed in \Cref{sec:app-sys-percep}.

\begin{wrapfigure}{r}{0.45\textwidth}
    \vspace{-1.75em}
    \begin{center}
        \includegraphics[width=1.0\linewidth]{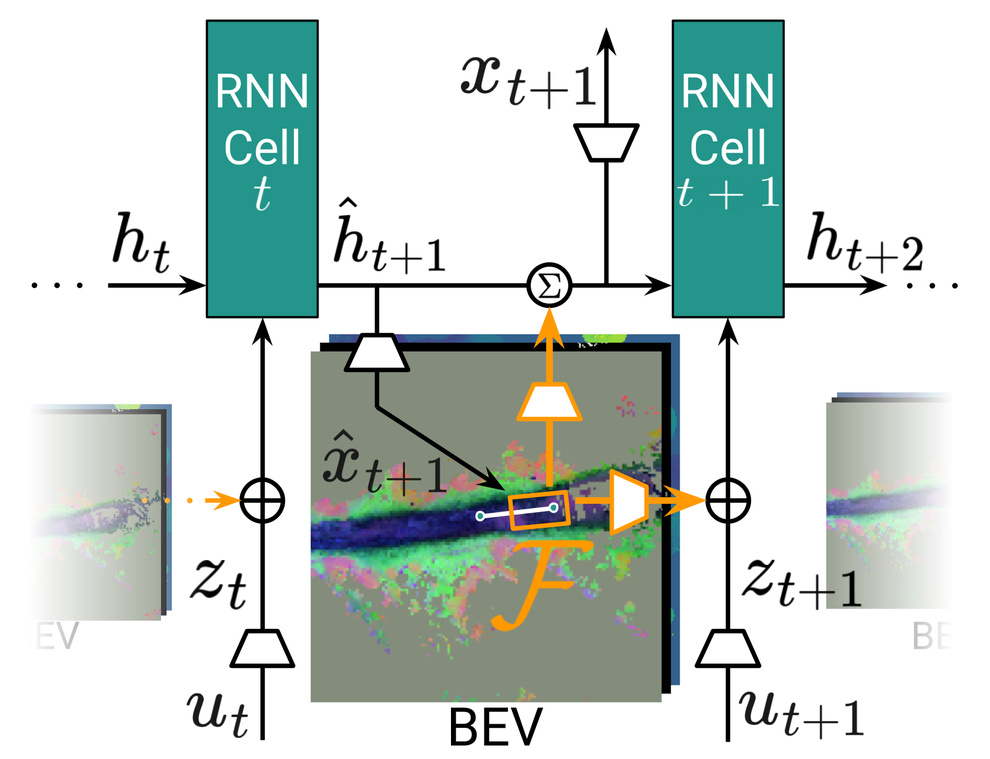}
    \end{center}
    \caption{Recurrent BEV Cells. At each step, the next predicted state determines a footprint over the BEV map. Sampled features are passed into the subsequent cell.}
    \label{fig:rec-bev-cell}
    \vspace{-1em}
\end{wrapfigure}

\paragraph{Recurrent BEV Cells}
Prior work concatenates a flattened BEV latent to the initial RNN hidden state. While this improves performance, it forces the network to learn spatial indexing from an unstructured vector. It discards the spatial prior that the most relevant BEV features at step $t+1$ are those the robot is about to physically interact with, from step $t$. We instead intercept the connection between consecutive RNN memory cells with a new recurrent BEV cell that injects this structured prior~(\Cref{fig:rec-bev-cell}) .

Concretely, let $\smash{\mathbf{B}} \in \mathbb{R}^{W\times H\times F}$ denote the BEV feature map and $\hat{\textbf{\rnnstate}}_{t+1}$ the hidden state at step $t+1$ from cell $t$. We decode the hidden state into a state difference $\Delta \textbf{\state}_t$ such
that~$\smash{\hat{\textbf{\state}}_{t+1}~=~\textbf{\state}_t~+~\encstate(\hat{\textbf{\rnnstate}}_{t+1})}$.
We define a set of tire track coordinates for our wheeled platform as the BEV footprint $\smash{\mathcal{F}\!\left(\hat{\textbf{\state}}_{t+1}\right) \subset \mathbb{R}^2}$ (Sec. \ref{sec:app-net-footprint}) and use it to sample over $\mathbf{B}$ to obtain the BEV feature context vector in eq. \ref{eq:bev-context-vec} where $\mathbf{B}[\cdot]$ denotes bilinear interpolation of the features in the BEV grid at the footprint points and $\encbev$ is a learned encoder. This context is used in two key manners. Most importantly, it applies a corrective update to the hidden state $\smash{\textbf{\rnnstate}_{t+1}=\hat{\textbf{\rnnstate}}_{t+1}+\textbf{\bevcontext}_{t+1}}$, such that each state prediction is informed by its current context vector at $t$ alongside its predicted context vector at $t+1$. Second, it augments the RNN cell's input latent action $\textbf{\rnninput}_{t+1}$, so the next state has its current BEV context.
For $\contextrnn$, we forgo the correction step and use the history of known states $\textbf{\state}_{t-\history:t}$ for query.
Formally, our structured perception representation is defined as
\begin{equation}
\label{eq:bev-context-vec}
    \mathbf{\bevcontext}_{t+1} \triangleq \encbev\!\left(\text{vec}\left(\mathbf{B}[\mathcal{F}(\hat{\mathbf{\state}}_{t+1})]\right)\right),
\end{equation}
\begin{equation}
    \textbf{\state}_{t+1} = \textbf{\state}_t+\encstate\left(\predrnn\left(\textbf{\rnnstate}_t,\textbf{\rnninput}_t,\textbf{\bevcontext}_{t}\right)+\textbf{\bevcontext}_{t+1}\right)
\end{equation}

This construction can be viewed as a variant of spatial attention, differing in how much of the sampling location is fixed by geometry rather than learned. Standard attention over the BEV map~\cite{li2022bevformer} assigns learned weights to every spatial location and recovers footprint locality through training. Learned offset approaches such as deformable attention \cite{zhu2020deformable} learn to predict where to sample. Our formulation reflects that of \citet{jaderberg_spatial_2015}, with a sampling grid fixed by construction. While our footprint corresponds to a wheeled platform, the sampling grid is the only platform-specific component; replacing $\mathcal{F}$ with the contact pattern of a quadrupedal, bipedal, or tracked system leaves the rest of the architecture unchanged.

\section{Benchmarking}
\label{sec:benchmark}

Dynamics models must be rigorously evaluated before deployment in the field. Aggregate metrics over large datasets $\dataset$ are dominated by easy, benign interactions and mask long-tail failure modes that determine real-world safety. We therefore mine the dataset for its hardest samples, enabling targeted evaluation on the state sequences models are most likely to mispredict. Intuitively, an identical action sequence, executed from an identical initial condition in an identical environment, will yield an identical trajectory. In reality however, environments are rarely identical, which motivates this work to learn environment terradynamics so as not to defer to epistemic uncertainty. Terradynamics are most challenging to model for off-road driving~\cite{rogers-marcovitz_continuous_2012, seegmiller_vehicle_2013, howard_optimal_2007}, so our goal is to formulate a dataset $\dataset_\text{hard} \subset \dataset$ that maximizes trajectory variance under similar actions, attempting to highlight where terradynamics are most prevalent.

\begin{figure}[t]
    \centering
    \includegraphics[width=1.0\linewidth]{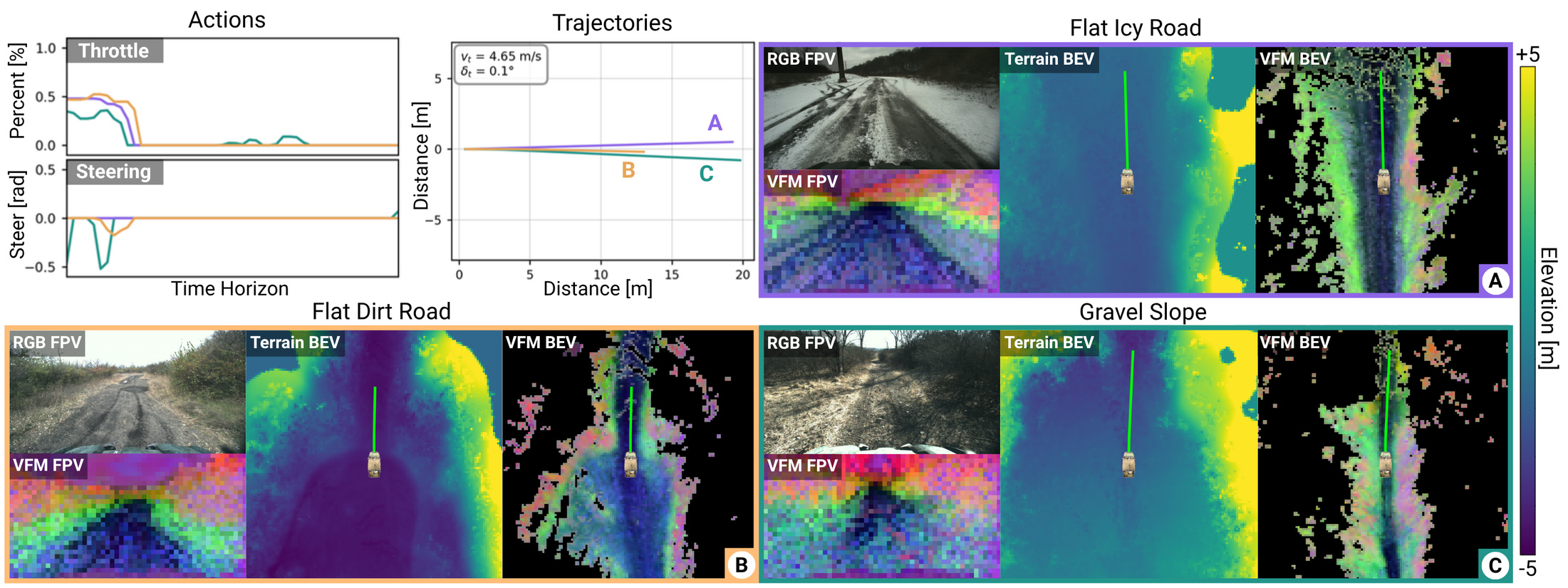}
    \caption{Similarly clustered actions yield different trajectories. Perception highlights how terradynamics are causing this difference. (C) travels further, going down a hill, as seen by the terrain map, (B) moves the least on flat dirt, while (A) travels further on flat, low-friction ice.}
    \label{fig:clustering-example}
\end{figure}

We define similarity between two candidates by the $\mathcal{L}_2$ distance between their normalized action sequences and initial conditions, connecting pairs that fall within a threshold and partitioning the resulting graph into cliques. The threshold is selected by maximizing the mean silhouette coefficient \cite{rousseeuw_silhouette_1987}. For a given group $\mathcal{G}$ of $N$ samples, the mean pairwise Hausdorff distance $\bar{d}_H$ is computed across all trajectories in the group as a measure of trajectory variance and difficulty score~(eq. \ref{eq:var}). \Cref{fig:clustering-example} shows an example group of similar candidates, where a difference in trajectories under similar actions can be attributed to downward slopes and ground friction of ice.

Sorting groups by this score ranks the dataset by difficulty. The benchmark defines $\dataset_\text{hard}$ as the union of the top $p$ percent of ranked groups, where $p$ controls how aggressively the long tail is targeted. The error metric itself is unchanged; only the distribution it is taken over differs (eq. \ref{eq:metric-hard}). A model may match its baselines on $\clusterobj_{\dataset}$ while $\clusterobj_{\text{hard}}$ diverges, revealing the long-tail failures that aggregate evaluation conflates with dominant benign data.
\begin{equation}
\label{eq:var}
    \bar{d}_H(\mathcal{G}) = \frac{1}{N(N-1)}\sum_{i=1}^{N}\sum_{j=1}^{N}
    d_H\!\left(\traj^i,\, \traj^j\right),\quad j\neq i
\end{equation}
\begin{equation}
\label{eq:metric-hard}
    \clusterobj_{\dataset} = \mathbb{E}_{\traj \sim \dataset}
    \left[\clusterobj(\hat{\traj}, \traj)\right]
    , \quad
    \clusterobj_{\text{hard}} = \mathbb{E}_{\traj \sim \dataset_{\text{hard}}}
    \left[\clusterobj(\hat{\traj}, \traj)\right]
\end{equation}

\section{Experiments}
\label{sec:results}
Using our benchmarking approach, we perform a series of online and offline evaluations across our dataset and real-world closed-loop control trials. We investigate our approach, following these key research questions:
\begin{itemize}
    \item[\rqfmt{RQ1}:] Do spatial perception representations and leveraging the spatial structure of the problem improve dynamics predictions?
    \item[\rqfmt{RQ2}:] Can our benchmarking approach mine and expose long-tail failure cases?
    \item[\rqfmt{RQ3}:] Does open-loop prediction error correlate to closed-loop control performance?
\end{itemize}

\paragraph{Training \& Evaluation}
We collect data at two test sites: \Gascola and \Renegade. The approach is trained on 79.8k samples (\unit{2.2}{hr}) of autonomous and teleoperated navigation around \Gascola. Importantly, this dataset is only collected in the spring, fall, and summer, with no snow data available. We present all results on a held-out dataset of 75.3k samples (\unit{2.1}{hr}) of autonomous and teleoperated navigation around \Gascola and \Renegade. \Renegade includes extreme hills not seen in the training set, and 34\% of the test set is composed of winter data with 6-9'' deep snow and thick ice which introduces additional tire forces and reduced friction.

Our approach uses a Long Short-Term Memory (LSTM) \cite{hochreiter1997lstm}, trained to predict \unit{5}{s} into the future with a $\Delta t$ of \unit{0.1}{s}, given a BEV map with geometry and visual semantic features, alongside \unit{5}{s} of prior historical state-action pairs. We train to minimize the Mean Squared Error (MSE) loss using the Adam optimizer \cite{kingma2017adam}. Our state reflects an extended kinematic state and is comprised of pose in $\text{SE}(2)$, speed, current steer, and pitch: $\smash{\textbf{\state}=\begin{bmatrix}x & y & \psi & v & \delta & \phi\end{bmatrix}}$.

\paragraph{Map Inputs}
Our BEV maps are \unit{100\times100}{m} at \unit{40}{cm} resolution. We train and evaluate with either sparse maps, which only use aggregate lidar mapping (Sec. \ref{sec:app-sys-percep}), or dense maps, which use a learned map inpainting network from \citet{triest2026trav}. Using both sparse and dense maps highlights how different BEV-conditioned methods leverage the BEV features.

\paragraph{Baselines}
We evaluate our approach against 4 baselines:
1) a single-track extended kinematic bicycle model with actuator modeling and parameters fit by least squares to ground-truth trajectories;
2) a state-only LSTM (State), with state history context, as a faithful recreation of~\citet{gibson_multi-step_2023} as a non-hybrid approach;
3) a FPV-conditioned LSTM (FPV), in which the FPV is encoded once at the start of propagation and combined with the state-history context;
4) a BEV-conditioned LSTM (\badbev), with the same, aforementioned initial encoding, similar to \citet{roth_learned_2025}.

\begin{wrapfigure}{r}{0.6\textwidth}
    \vspace{-0.5cm}
    \centering
    \includegraphics[width=1.0\linewidth]{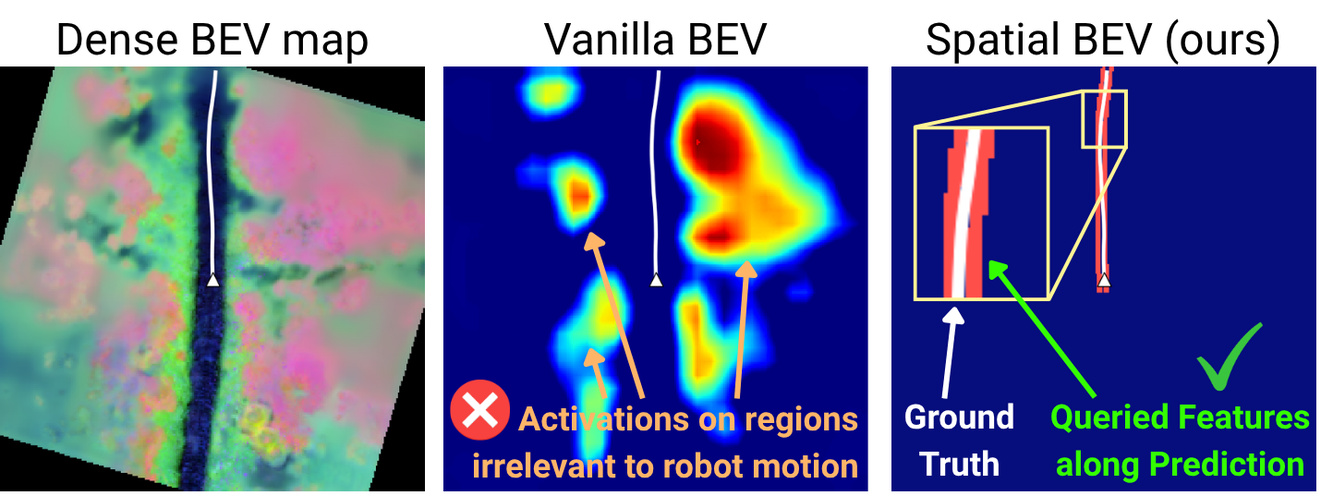}
    \caption{\badbev activations and our queries with reference BEV map and ground truth trajectory (white). \badbev's activations fall on irrelevant regions, contrary to our queries along our prediction near the ground truth.}
    \label{fig:actmap-qual}
    \vspace{-0.5cm}
\end{wrapfigure}

\begin{figure}[t]
    \centering
    \includegraphics[width=1.0\linewidth]{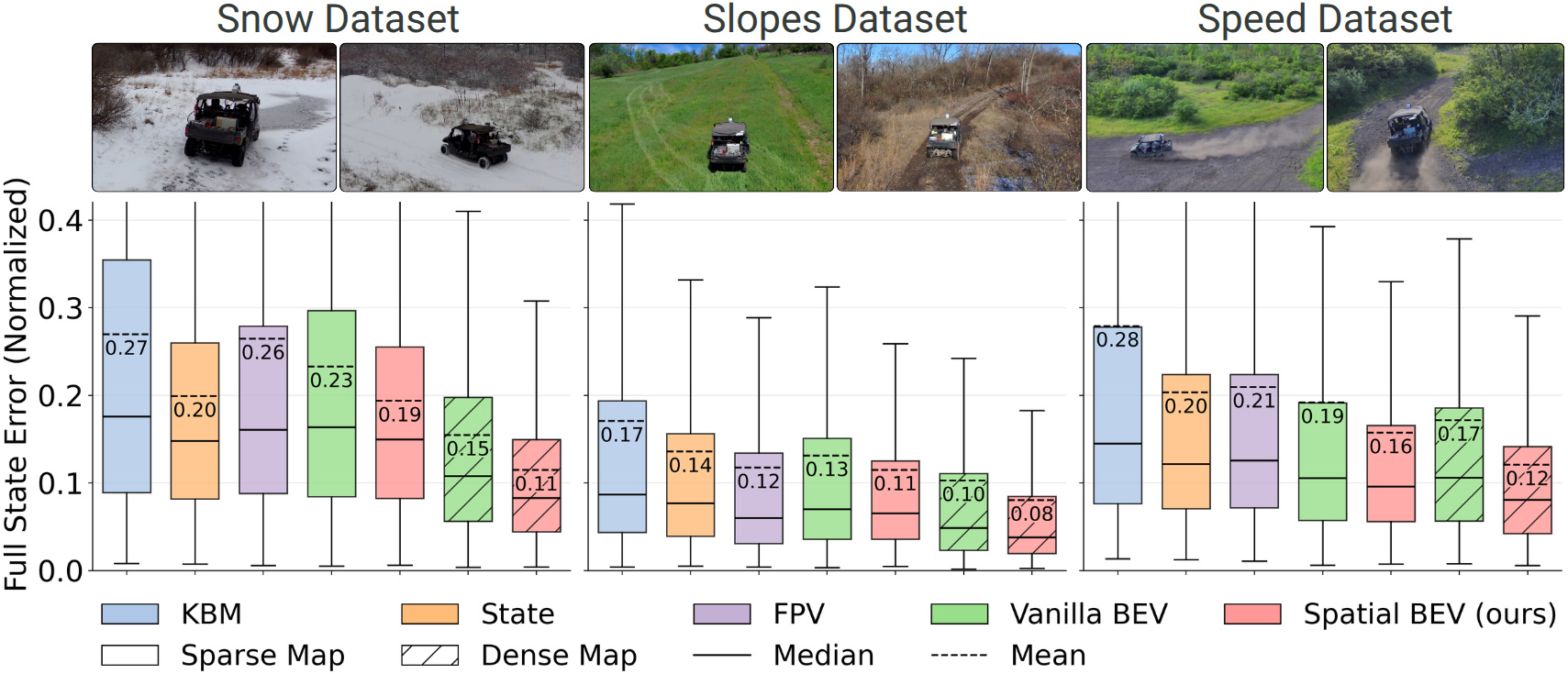}
    \caption{Offline full state prediction error distributions across 3 datasets. The Snow dataset is most challenging for all methods. Dense maps reduce error mean and distribution variance. Across all datasets, our method outperforms the baselines.}
    \label{fig:offline-quant}
    \vspace{1em}
    \centering
    \small
    \resizebox{\linewidth}{!}{
    \begin{tabular}{lccccccccc}
        \toprule
         & & \multicolumn{2}{c}{Snow} & \multicolumn{2}{c}{Slopes} & \multicolumn{2}{c}{Speed} & \multicolumn{2}{c}{Full} \\
        \cmidrule(lr){3-4} \cmidrule(lr){5-6} \cmidrule(lr){7-8} \cmidrule(lr){9-10}
        Method & Map & MSE [$\text{m}^2$] $\downarrow$ & P95 & MSE [$\text{m}^2$] $\downarrow$ & P95 & MSE [$\text{m}^2$] $\downarrow$ & P95 & MSE [$\text{m}^2$] $\downarrow$ & P95 \\
        \midrule
        KBM & -- & 2.68 $\pm$ 4.23 & 8.92 & 2.31 $\pm$ 4.71 & 8.00 & 2.83 $\pm$ 6.52 & 10.24 & 2.10 $\pm$ 4.24 & 7.20 \\
        State & -- & 2.54 $\pm$ 2.52 & 7.62 & 2.20 $\pm$ 3.15 & 7.48 & 1.91 $\pm$ 3.69 & 7.39 & 1.99 $\pm$ 2.75 & 6.71 \\
        FPV & -- & 2.34 $\pm$ 2.37 & 6.93 & \underline{1.69 $\pm$ 2.95} & \underline{5.82} & 2.12 $\pm$ 4.55 & 7.82 & 1.77 $\pm$ 2.83 & 6.00 \\
        Vanilla BEV & \sparsebox & 2.49 $\pm$ 2.89 & 8.12 & 1.98 $\pm$ 3.39 & 7.20 & 1.59 $\pm$ 3.47 & 6.33 & 1.73 $\pm$ 2.87 & 6.66 \\
        Vanilla BEV & \densebox & \underline{2.10 $\pm$ 2.44} & \underline{6.76} & 1.97 $\pm$ 3.28 & 7.70 & \underline{1.41 $\pm$ 2.93} & \underline{5.17} & \underline{1.46 $\pm$ 2.50} & \underline{5.63} \\
        \hline\noalign{\vspace{2pt}}
        Spatial BEV (ours) & \sparsebox & 2.45 $\pm$ 2.66 & 8.06 & 1.78 $\pm$ 2.51 & 6.87 & 1.50 $\pm$ 3.90 & 5.42 & 1.70 $\pm$ 2.68 & 6.26 \\
        Spatial BEV (ours) & \densebox & \textbf{1.74 $\pm$ 1.98} & \textbf{5.55} & \textbf{1.51 $\pm$ 2.44} & \textbf{5.26} & \textbf{1.13 $\pm$ 2.95} & \textbf{3.58} & \textbf{1.21 $\pm$ 2.14} & \textbf{4.41} \\
        \bottomrule
    \end{tabular}
    }
    \captionof{table}{Offline position prediction MSE across three datasets, alongside the full test set. BEV maps are sparse \sparsebox~or dense \densebox. Our spatial approach outperforms its \badbev counterpart with sparse and dense maps. With dense maps, our approach outperforms all baselines.}
    \vspace{-0.5em}
    \label{tab:offline-quant}
\end{figure}

\subsection{Effect of Structured Perception Representations}
\label{sec:exp-offline}
To answer \rqfmt{RQ1} and determine the benefit of structured perception representations, we evaluate our approach on the open-loop prediction training task, provide a qualitative analysis on the benefit of our representation, and ablate our architecture design.

\paragraph{Open-loop Prediction}
\label{sec:rq1-preds}
We first evaluate our approach on the training task of open-loop prediction with no closed-loop actuation using MSE on the full and position state spaces. We evaluate on three sub-datasets: Snow, Slopes, and Speed, defined in~\Cref{sec:app-dataset}. The benefit of our structured representation is evident in \Cref{fig:offline-quant} and \Cref{tab:offline-quant}, where our approach with dense maps reduces prediction error relative to the strongest baseline of each sub-dataset by 17.1\% on Snow, 8.5\% on Slopes, 19.9\% on Speed, and 17.1\% on Full. Even with sparse maps, our approach outperforms baselines on Speed and Full datasets; however, FPV has lower error on Snow and Slopes. This is expected, since LiDAR coverage degrades on slopes, leaving sparse BEV maps least informative. Finally, comparing against our \badbev counterpart isolates the effect of our representation. Holding the map input fixed, we improve by 4.6\% with sparse maps and 19.4\% with dense maps on average, highlighting the benefit of our representation and addressing \rqfmt{RQ1}. Qualitative model rollouts can be found in~\Cref{sec:offline-qual}.

\paragraph{Qualitative Representation Analysis}
\label{sec:rq1-qual}
Intuitively, our structured representation queries local environment features which improves predictions, but we seek to explain \textit{why} this intuition holds. We visualize the activation maps of \badbev and our explicit approach in~\Cref{fig:actmap-qual} and see \badbev's activations fall on irrelevant map regions, whereas our queries fall along our prediction, which closely follows the robot's true path. This affirms our intuition and illustrates the benefit of our representation. More examples and activation map computation details in~\Cref{sec:app-net-gradcam}.

\paragraph{Architectural Decisions}
\label{sec:rq1-ablations}

\begin{wraptable}{r}{0.45\textwidth}
    \centering
    \resizebox{\linewidth}{!}{
    \begin{tabular}{@{}c|@{\hspace{4pt}}lccccc}
        \toprule
        \multicolumn{2}{l}{} && \multicolumn{4}{c}{MSE [$\text{m}^2$] $\downarrow$} \\
        \cmidrule(lr){4-7}
        \multicolumn{2}{l}{Ablations} & Map & Snow & Slopes & Speed & Full \\
        \midrule
        \multirow{4}{*}{\rotatebox[origin=c]{90}{BEV Cell}}
        & Non-corr. & \sparsebox & 3.54 & 2.18 & 1.89 & 2.19 \\
        & Corr. & \sparsebox & 2.45 & 1.78 & 1.50 & 1.70 \\
        & Non-corr. & \densebox & \underline{1.98} & \underline{1.56} & \textbf{1.11} & \underline{1.27} \\
        & Corr. & \densebox & \textbf{1.74} & \textbf{1.51} & \underline{1.13} & \textbf{1.21} \\
        \midrule
        \multirow{4}{*}{\rotatebox[origin=c]{90}{Footprint}}
        & 4-point & \densebox & \underline{1.79} & 1.88 & 1.49 & \underline{1.41} \\
        & Tracks & \densebox & \textbf{1.74} & \textbf{1.51} & \textbf{1.13} & \textbf{1.21} \\
        & Full body & \densebox & 2.05 & \underline{1.68} & \underline{1.35} & 1.50 \\
        & Lookahead & \densebox & 1.87 & 2.03 & 1.54 & 1.55 \\
        \bottomrule
    \end{tabular}
    }
    \caption{Ablations of the correction step (top) and footprint (bottom) on all datasets.}
    \label{tab:ablations}
\end{wraptable}

We motivate our architecture's correction step and footprint via ablations in \Cref{tab:ablations}. The recurrent BEV cell correction reduces error on the full dataset by 22.4\% on sparse maps and 4.7\% on dense maps, with up to 30.8\% and 12.1\% on snow, and reduces error in all but one setting of dataset and map input.
We attribute the smaller dense map gain to the richer input leaving less to correct. Next, we compare footprints described in~\Cref{sec:app-net-footprint-ablation}. The Tracks footprint reduces error on the full dataset by 14.2\% over the next best footprint and is the best across all sub-datasets. 
Compared to 4-point, Tracks densely cover the tire path, which matters most at speed (24.2\% improvement), where the robot moves further than the map resolution per step. However, the coverage must be balanced; the Full body and Lookahead footprints query cells away from the tire path which dilutes the relevant information.

\subsection{Mining Long-tail Failures}
\begin{wrapfigure}{r}{0.4\textwidth}
    \vspace{-3em}
    \centering
    \includegraphics[width=1.0\linewidth]{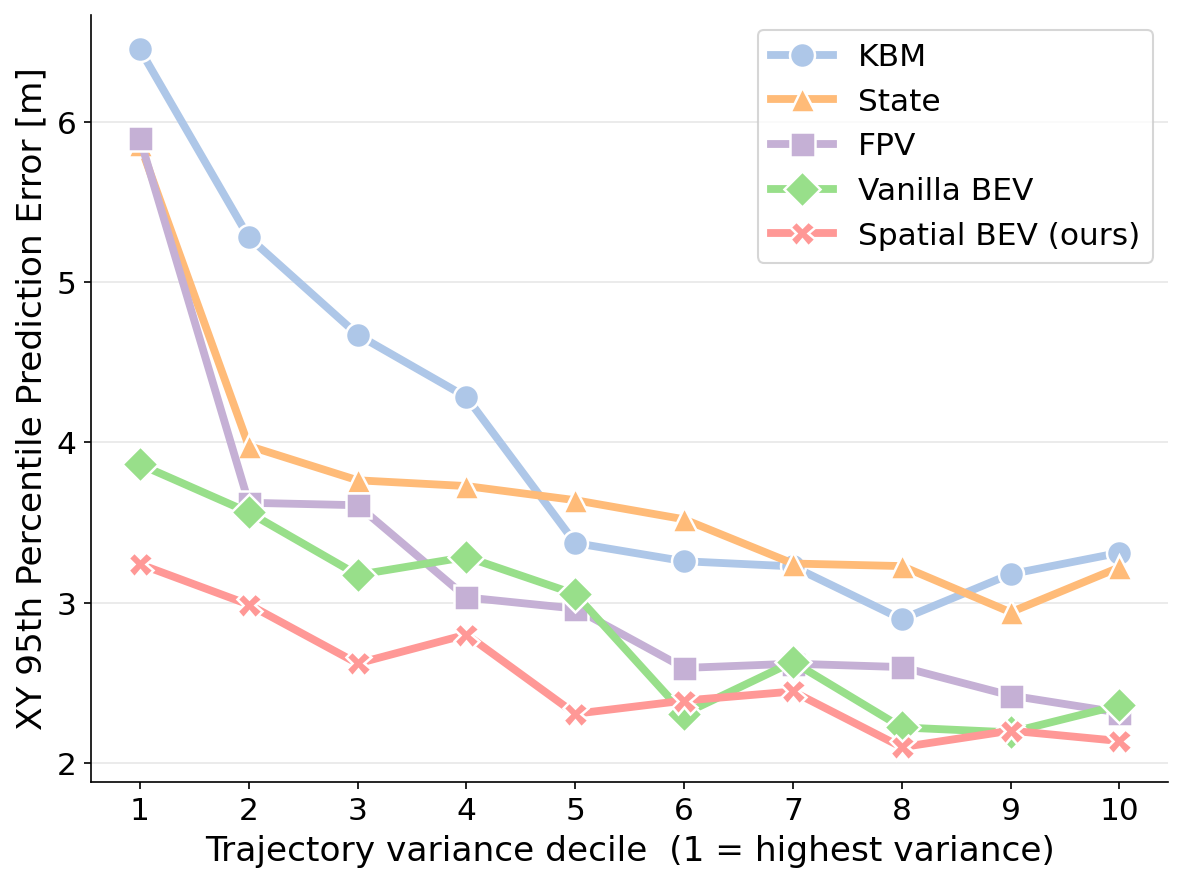}
    \caption{Evaluation on the highest to lowest variance deciles of the clustered benchmark. Model performance increases with lower variance deciles.}
    \label{fig:clustering-variance}
    \vspace{-0.5cm}
\end{wrapfigure}

We want to validate our benchmarking method by confirming if we are able to mine the most challenging data $\dataset_\text{hard}$ from a broader dataset $\dataset$, answering \rqfmt{RQ2}. To do this, we must determine if we can identify the most challenging data. We partition our dataset sorted by trajectory variance into 10 equal partitions or deciles, such that the \nth{1} decile should have the highest variance and \nth{10} decile has lowest. We evaluate all methods on each decile in \Cref{fig:clustering-variance} and find that all models' worst-case prediction error is highest on the highest variance decile and lowest on the lowest variance decile  with a clear downwards trend. This demonstrates how our benchmark can identify the highest variance runs and how this correlates to dynamics modeling difficulty. Moreover, on the most challenging partition, we see the FPV and State baselines have much higher error, highlighting the importance of evaluating on the long-tail.

\subsection{Closed-loop Control on Hardware}
\label{sec:online-results}
We further evaluate our approach on hardware and investigate how prediction error is correlated with closed-loop performance to answer \rqfmt{RQ3}. We deploy our solutions on a full-sized Yamaha Viking ATV, retrofitted with autonomy sensors and actuators~\cite{Mai-2020, sivaprakasam2024tartandrive2}.

\begin{figure}[t]
    \centering
    \includegraphics[width=1.0\linewidth]{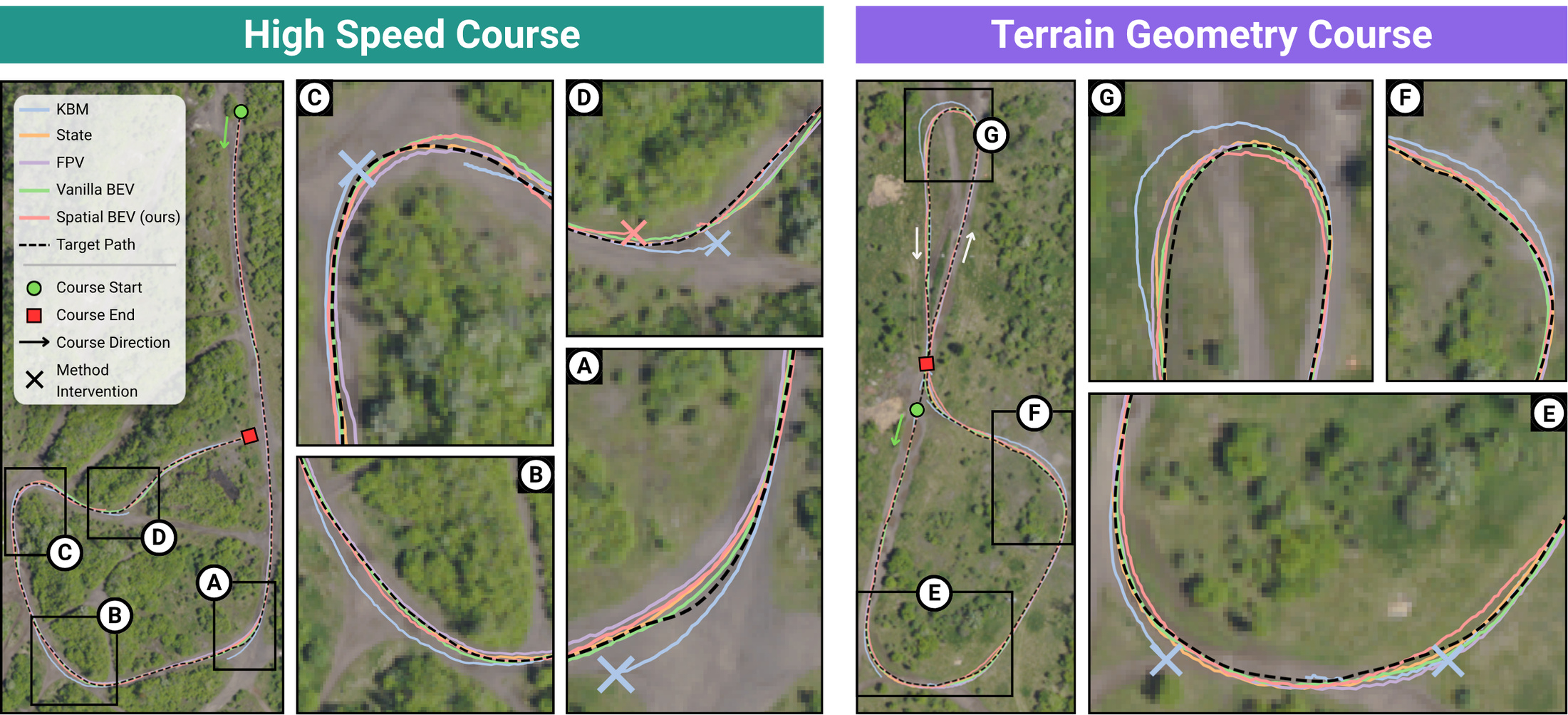}
    \caption{Experiments for hardware evaluation on two challenging courses. Significant understeer can be seen by the KBM baseline leading to numerous interventions (A,D,E). Additionally all learned methods, including ours, experience cases of understeer like in the tight turn in C or G.}
    \label{fig:online-qual}
    \vspace{1em}
    \centering
    \begin{tabular}{lrrrrrrrrrr}
    \toprule
     & \multicolumn{5}{c}{High Speed Course (1 trial)} & \multicolumn{5}{c}{Geometry Course (4 trials)} \\
    \cmidrule(lr){2-6} \cmidrule(lr){7-11}
    Method & \shortstack{\# Intv\\avg} & \shortstack{vel.\\avg} & \shortstack{vel.\\max} & \shortstack{CTE\\avg} & \shortstack{CTE\\max} & \shortstack{\# Intv\\avg} & \shortstack{vel.\\avg} & \shortstack{vel.\\max} & \shortstack{CTE\\avg} & \shortstack{CTE\\max} \\
    \midrule
    KBM & 3.00 & 7.75 & 11.20 & 0.73 & 5.04 & 3.75 & 6.13 & 9.01 & 0.81 & 5.51 \\
    State & \textbf{0.00} & 8.39 & 10.96 & \textbf{0.42} & \textbf{2.29} & \textbf{0.00} & 6.80 & 8.68 & 0.45 & 3.35 \\
    FPV & \textbf{0.00} & 7.85 & 10.98 & 0.80 & 3.41 & 0.75 & 6.50 & 8.90 & 0.65 & 4.90 \\
    Vanilla BEV & \textbf{0.00} & 8.44 & 10.94 & 0.47 & 3.26 & \textbf{0.00} & 6.94 & 8.60 & \textbf{0.43} & 3.67 \\
    \midrule
    Spatial BEV (ours)\hspace{-1.5em} & 1.00 & 8.02 & 10.41 & 0.51 & 2.62 & \textbf{0.00} & 6.58 & 8.66 & 0.60 & \textbf{2.74} \\
    \bottomrule
    \end{tabular}
    \addtocounter{table}{1} 
    \captionof{table}{Hardware deployment results on two challenging courses.}
    \label{tab:online-quant}
\end{figure}

For closed-loop evaluation, we use Model Predictive Path Integral (MPPI) control~\cite{williams_aggressive_2016, williams_information_2017} with $825$ particles at \unit{10}{Hz}, whose actions are executed by a \unit{30}{Hz} low-level controller. MPPI follows a reference line subject to lateral acceleration constraints (Sec. \ref{sec:app-sys-control}). We measure cross-track error (CTE) of all methods and number of interventions as a light heuristic. We were unable to run BEV-based methods with dense maps.
We evaluate on two courses at \Gascola, shown in~\Cref{fig:online-qual}:~1) A high speed, \unit{739}{m} course over flat gravel and dirt terrain with a target speed of \unit{10}{m/s} (1 trial); 2) A \unit{524}{m} course over challenging terrain, such as steep slopes and banked turns, with a target speed of \unit{8}{m/s} (4 trials).

\begin{wrapfigure}{l}{0.4\textwidth}
\small
\centering
\resizebox{\linewidth}{!}{
\begin{tabular}{lccc}
    \toprule
    Method & Map & MSE [$\text{m}^2$] $\downarrow$ & P95 $\downarrow$ \\
    \midrule
    KBM & -- & 2.91 $\pm$ 3.18 & 10.20 \\
    State & -- &  1.60 $\pm$ 1.52 & 5.00 \\
    FPV & -- &  1.78 $\pm$ 1.61 & 5.38 \\
    Vanilla BEV & \sparsebox &  1.35 $\pm$ 1.53 & 4.86 \\
    Vanilla BEV & \densebox & \underline{1.26 $\pm$ 1.24} & \underline{3.97} \\
    \midrule
    Spatial BEV & \sparsebox & 1.57 $\pm$ 1.34 & 4.43 \\
    Spatial BEV & \densebox & \textbf{1.05 $\pm$ 0.83} & \textbf{2.82} \\
    \bottomrule
\end{tabular}}
\addtocounter{table}{-2}
\captionof{table}{Offline prediction results on closed-loop experiment data}
\label{tab:online-offline-quant}
\addtocounter{table}{1}
\end{wrapfigure}

While all models can track straight lines well at speed, the turns prove to be the most challenging to track. The KBM baseline consistently understeers in low and high curvature turns, resulting in at least 3.00 average interventions per trial across both courses. From Table \ref{tab:online-quant}, our approach matches closed-loop performance of the other learned baselines, even with sparse maps. While we demonstrate competitive average CTE on the high speed and geometry courses, our approach has the lowest worst-case CTE for the geometry course, aligning with the open-loop results of reducing worst-case performance. To help isolate the effect of the controller, we evaluate prediction error on the experiment data in Table~\ref{tab:online-offline-quant}. We find while our CTE is closer to the baseline performance with the closed-loop control, our prediction error is still 16.6\% better than the best baseline on average.  While one intervention did happen for our method, this was diagnosed in~\Cref{sec:app-sys-mppi-fail} as an MPPI-tracking issue instead of a dynamics modeling inaccuracy, but it is listed here for transparency. This answers \rqfmt{RQ3} and shows open-loop and closed-loop performance are not always correlated due to the entanglement of the controller.

\section{Limitations}
First, although our approach performs well on held-out datasets and environments, online adaptation could further improve robustness and generalization \cite{lupu_magic-vfm_2024, levy_learning_2024, oconnell_neural-fly_2022, sivaprakasam2025salon, liu2025locoformer}. Second, training on real-world data avoids a sim-to-real gap but is expensive, sample-inefficient, and vulnerable to hardware failures; combining simulation and real data is a natural next step \cite{xiao_anycar_2024, levy_learning_2024}. 
Third, map querying and interpolation add computational cost. \badbev rolls out 825 particles in \unit{12.9}{ms} versus \unit{28.3}{ms} for ours, rising to \unit{40}{ms} and \unit{82}{ms}, respectively, with all components running. While sufficient for \unit{10}{Hz} control, this leaves little room for more samples or a longer horizon.
Finally, a better line tracking algorithm could be used for future testing, given the false intervention of our approach.

\section{Conclusion}
This paper presents structured perception representations for learned dynamics models for off-road terrain. We demonstrate how off-road terrain, represented through geometric or visual semantics, should not be left as unmodeled epistemic uncertainty but rather a core component to the dynamics model. Through extensive offline evaluation and online hardware results, we demonstrate the necessity of including perception representations in dynamics models and how additional spatial structure further improves dynamics modeling on high speed ($\geq$10m/s) and geometrically challenging courses.



\clearpage
\acknowledgments{We thank Geethika Hemkumar and Saksham Sharma for their extensive support with field testing for the submission and rebuttal. We also thank Jason Gibson and Anoushka Alavilli for insightful discussions during the beginning of this work. Micah Nye is supported by the Rales Fellowship. This work was supported by DSO National Laboratories by grant \#DSOCO25020 RPS 22. This work used Bridges-2 at Pittsburgh Supercomputing Center through allocation cis220039p from the Advanced Cyberinfrastructure Coordination Ecosystem: Services \& Support (ACCESS) program, which is supported by National Science Foundation grants \#2138259, \#2138286, \#2138307, \#2137603, and \#2138296.}


\bibliography{ref}  

\clearpage
\appendix
\section{Additional Experiments}
We further evaluate our approach qualitatively and quantitatively. Specifically, we investigate the qualitative offline results of our approach, compared to other methods and provide more insight on how our method's proper use of structured perception representations improve its forward dynamics prediction (Sec. \ref{sec:offline-qual}). Next, we evaluate our approach against more methods, which do not use the RNN-based dynamics backbone and instead use Transformers (Sec. \ref{sec:offline-quant-extended}). Finally, we provide insight to our perception representation via seed ablations (Sec. \ref{sec:spatial-representation-ablation}).

\subsection{Open-loop Qualitative Analysis}
\label{sec:offline-qual}
\begin{figure}[t]
    \centering
    \includegraphics[width=1.0\linewidth]{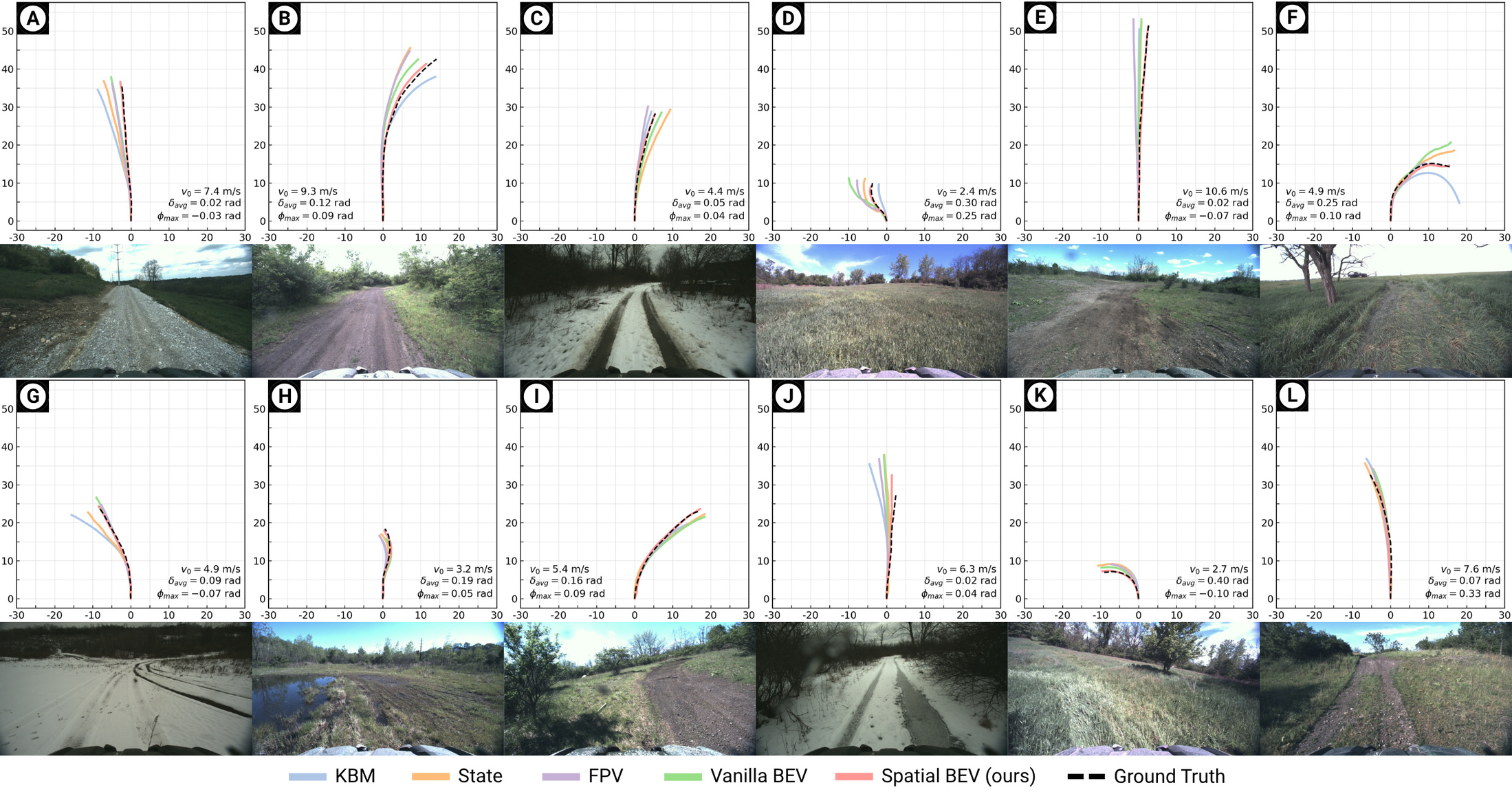}
    \caption{Offline qualitative results highlighting interesting prediction examples and environments.}
    \label{fig:offline-qual}
\end{figure}

We qualitatively evaluate the success and failure cases of our approach by analyzing the prediction results in Figure \ref{fig:offline-qual}. Across a diverse set of environments and motion profiles, including snow (C), ice (J), marshes (H), uphill grades (D,K,L), our method consistently produces accurate trajectory predictions at low, medium, and high speeds, spanning straight segments (A,E), slight and large turns (G,I), sharp turns (F,K), and slalom maneuvers (D,H). Notably in (G), all perception-based methods perform comparably well on a slight snowy turn, highlighting conditions where appearance-based cues alone provide enough context.

In challenging uphill scenarios (L), all methods overshoot due to poor terrain understanding ahead, while our approach remains closest to the ground truth with only a slight undershoot, reflecting our model's improved terrain awareness. The most difficult case arises in (J), where ice on the right breaks under the robot's weight, causing the vehicle to slow and its tires to get lodged in a newly formed deep rut. All methods overshoot, predicting trajectories to the left of the actual path, since the rut constrains the robot's motion. No method is able to predict this phenomenon, regardless of representation.

\subsection{Extended Open-loop Results}
\label{sec:offline-quant-extended}
While we restricted our evaluation to an LSTM-based ablation for brevity, we provide further offline quantitative results on popular alternatives in sequential prediction, such as Transformer-based architectures \cite{vaswani2017transformer}. We choose to evaluate against the following baselines: 
\begin{enumerate}
    \item State-only Transformer (State TF), which tokenizes the history of state-action pairs alongside future actions. We architect it as a Transformer encoder, with a causal attention mask on the state-action pairs.
    \item BEV Transformer (BEV TF), which additionally tokenizes a BEV map for perceptual conditioning. We tokenize the BEV map by encoding each $5\times5$ cell patch into one token. We architect this as a Transformer decoder, with a causal attention mask on the state-action pairs and full-attention between BEV and state-action pairs, so each state or action token is able to attend to the most relevant BEV patches.
    \item AnyCar \cite{xiao_anycar_2024}: a state-of-the-art, state-based Transformer for dynamics prediction. We train from scratch on the same data all LSTM's were trained on in Section \ref{sec:results}. Notably, we train it to predict 5s sequences at 10Hz, whereas, \citet{xiao_anycar_2024} predict 1s in advance at 50Hz. Likewise the history differs in this manner.
\end{enumerate}

Table \ref{tab:offline-quant-extended} extends the main results to include these Transformer baselines. The Transformer baselines achieve low prediction error on the Snow and Slopes datasets but degrade on the Speed dataset. The BEV Transformer performs worst across nearly all methods and datasets, suggesting that explicit sampling grids are a more effective spatial attention mechanism. Granted, its architecture is relatively simple and could be improved in future work.

The advantage of our structured representation is most apparent on the Speed dataset, where our method achieves the best average and worst-case performance on trajectories with large displacements from the starting position. This supports our conjecture that explicit spatial structure helps the dynamics model attend to the most relevant environmental features.

\begin{table}
\centering
\small
\begin{tabular}{lccccccccc}
    \toprule
     & & \multicolumn{2}{c}{Snow} & \multicolumn{2}{c}{Slopes} & \multicolumn{2}{c}{Speed} & \multicolumn{2}{c}{Full} \\
    \cmidrule(lr){3-4} \cmidrule(lr){5-6} \cmidrule(lr){7-8} \cmidrule(lr){9-10}
    Method & Map & MSE $\downarrow$ & P95 $\downarrow$ & MSE $\downarrow$ & P95 $\downarrow$ & MSE $\downarrow$ & P95 $\downarrow$ & MSE $\downarrow$ & P95 $\downarrow$ \\
    \midrule
    KBM & -- & 2.68 & 8.92 & 2.31 & 8.00 & 2.83 & 10.24 & 2.10 & 7.20 \\
    State & -- & 2.54 & 7.62 & 2.20 & 7.48 & 1.91 & 7.39 & 1.99 & 6.71 \\
    FPV & -- & 2.34 & 6.93 & \underline{1.69} & \underline{5.82} & 2.12 & 7.82 & 1.77 & 6.00 \\
    Vanilla BEV & \sparsebox & 2.49 & 8.12 & 1.98 & 7.20 & 1.59 & 6.33 & 1.73 & 6.66 \\
    Vanilla BEV & \densebox & \underline{2.10} & \underline{6.76} & 1.97 & 7.70 & \underline{1.41} & \underline{5.17} & \underline{1.46} & \underline{5.63} \\
    \hline\noalign{\vspace{2pt}}
    State TF & -- & 2.40 & 7.53 & \underline{1.69} & 6.34 & 1.83 & 7.24 & 1.75 & 6.21 \\
    BEV TF & \sparsebox & 3.61 & 10.29 & 2.06 & 7.83 & 3.46 & 12.54 & 2.37 & 8.56 \\
    AnyCar~\cite{xiao_anycar_2024} & -- & 2.42 & 7.92 & 1.92 & 7.89 & 1.66 & 6.40 & 1.70 & 6.51 \\
    \hline\noalign{\vspace{2pt}}
    Spatial BEV (ours) & \sparsebox & 2.45 & 8.06 & 1.78 & 6.87 & 1.50 & 5.42 & 1.70 & 6.26 \\
    Spatial BEV (ours) & \densebox & \textbf{1.74} & \textbf{5.55} & \textbf{1.51} & \textbf{5.26} & \textbf{1.13} & \textbf{3.58} & \textbf{1.21} & \textbf{4.41} \\
    \bottomrule
\end{tabular}
\captionof{table}{Extended offline position prediction results, including Transformer-based evaluations, evaluated across 3 datasets, alongside the full test set. Standard deviation removed for width-fitting.}
\label{tab:offline-quant-extended}
\end{table}

\subsection{Seed Ablations}
\label{sec:spatial-representation-ablation}
While offline results and ablations show success of the architecture, we perform a seed ablation. Over four random seeds, we see a prediction error standard deviation of \unit{0.07}{$\text{m}^2$}.

\section{Dataset Details}
\label{sec:app-dataset}
We collect our dataset using human expert demonstrations or prior autonomous navigation on an off-road all-terrain vehicle in the real world across two test sites in all four seasons. The entire dataset encompasses \unit{16.3}{hr} of driving data, prior to post-process filtering and sub-dataset partitioning. This dataset has top speeds of \unit{13}{m/s} and max pitch of 19.1$^\circ$ (32.2\% grade) with further dataset statistics detailed in Table \ref{tab:dataset-stats}. The dataset includes terrain ground cover such as grass, loose gravel, snow, ice, mud, and standing water, as well as geometry like steep slopes, banks, ruts, and undulating terrain. These features provide a diversity in geometric complexity, friction properties, and terrain deformability.

Using this large dataset, we split it into our train and test datasets as mentioned in Section \ref{sec:results}. Our test dataset is broken up into Snow, Slopes, and Speed sub-datasets for evaluation. These sub-datasets are defined in the following ways:
\begin{enumerate}
    \item Snow: Our dataset was manually filtered using run log metadata to find the runs with snow. Having snow was defined as having snow completely cover the ground.
    \item Slopes: We filtered our dataset to ensure, for a given trajectory $\traj$, the maximum pitch experienced is at at least 0.1 radians (5.8$^\circ$).
    \item Speed: We filtered out data to ensure, for a given trajectory $\traj$, the average speed is at least \unit{6.0}{m/s}.
\end{enumerate}

Post-process filtering filters negative speed to remove reverse motions, samples with large odometry covariance, and samples with braking, since brake is not in the robot's action space (Sec. \ref{sec:hardware}).

\begin{table}[]
    \centering
    \footnotesize
    \setlength{\tabcolsep}{3pt}
    \begin{tabular}{lrrrrrrr}
    \toprule
      & Mean & Std & Min & 5\% & Median & 95\% & Max \\
    \midrule
     Forward vel. (m/s) & 4.66 & 1.87 & -0.12 & 1.92 & 4.40 & 8.27 & 13.01 \\
     Lateral vel. (m/s) & 0.02 & 0.27 & -1.61 & -0.37 & 0.01 & 0.44 & 3.89 \\
     Slip (deg) & 0.27 & 4.41 & -38.43 & -6.10 & 0.13 & 7.24 & 51.23 \\
     Yaw Rate (rad/s) & -0.00 & 0.14 & -0.89 & -0.24 & -0.00 & 0.24 & 0.84 \\
     Pitch (deg) & -0.07 & 3.84 & -18.92 & -7.34 & 0.12 & 6.71 & 19.10 \\
     Roll (deg) & 0.22 & 3.31 & -18.85 & -5.28 & 0.37 & 5.22 & 20.07 \\
    \bottomrule
    \end{tabular}
    \caption{Statistics of entire dataset, encompassing test and train datasets}
    \label{tab:dataset-stats}
\end{table}

\section{System Implementation \& Analysis}
\subsection{Software}
\subsubsection{Perception}\label{sec:app-sys-percep}
In order to obtain inputs for our learned model, we leverage a GPU-accelerated local mapping package that produces geometric-semantic BEV maps from camera and LiDAR data. This
process is shown in~\Cref{fig:percep-pipeline} and accomplished via the following procedure:

\begin{figure}
    \centering
    \includegraphics[width=1.0\linewidth]{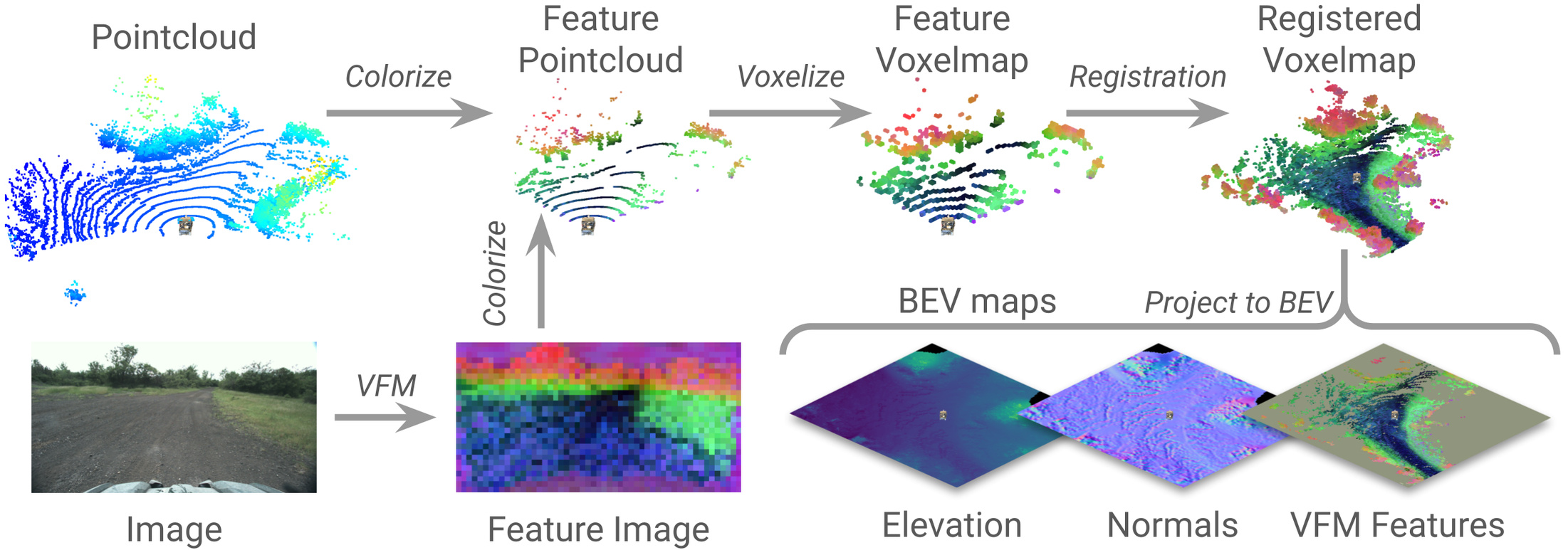}
    \caption{Perception and mapping flow diagram}
    \label{fig:percep-pipeline}
\end{figure}

\begin{enumerate}
    \item Salient visual features are extracted from the front-facing camera images using c-Radiov3 \cite{ranzinger2023radio} with a SIGLIP2 adaptor \cite{zhai2023sigmoid} and NACLIP attention \cite{hajimiri2025pay} and are then passed through a domain-specific principal component analysis (in a similar fashion to RayFronts \cite{alama2025rayfronts}). This results in a $53 \times 28 \times 32$ ``feature image''.
    \item This feature image is then lifted into 3D via pointcloud colorization, in which the LiDAR point cloud is projected into the feature image frame. Points that lie within the image frame inherit the feature from the pixel that they fall into (Eq \ref{eq:point_projection}), resulting in a ``feature point cloud''.
    \item The feature point cloud is then transformed into a fixed frame (via an on-board LiDAR-interial odometry algorithm \cite{zhao2021super}), voxelized, and aggregated into a scrolling voxel grid centered at the robot's current position.
    \item Finally, we use the voxel grid to estimate the underlying terrain surface of the robot's local environment using a Markov random field \cite{wellington2005interacting, maturana2022semantic}. The geometric features in the input BEV map such as step height and slope, are computed using the voxel grid geometry and the terrain surface. VFM features are computed as the average feature of all voxels within a threshold (roughly the ATV height = \unit{1.7}{m}) of the terrain surface.
\end{enumerate}

\begin{equation}
    \label{eq:point_projection}
    [u', v', w]^\intercal = \mathbf{K} [\mathbf{R} | \mathbf{t}] [x, y, z, 1]^\intercal, \quad u = \frac{u'}{w}, \quad v = \frac{v'}{w}
\end{equation}

This mapping system is run using the on-board GPU (Sec. \ref{sec:hardware}) at \unit{5}{Hz} to produce BEV maps for training and inference. We apply minor post-processing on the BEV maps before inference:
\begin{enumerate}
    \item To avoid the need for the network to learn BEV orientation and feature rasterizing, we transform the BEV map into an ego-aligned frame such that forward is always facing right. We do this by creating a grid of ego-frame coordinates, centered at $(0,0)$ in ego space. Next, we rotate this grid into odom frame. We then sample the odom BEV map at the ego coordinates in odom frame.
    \item We modify the features to limit local terrain elevation from $[\unit{-10, 10}{m}]$ and we rotate geometric data, such as surface normals, to be aligned in ego frame.
\end{enumerate}

\subsubsection{Control}
\label{sec:app-sys-control}
Our path tracking results used MPPI \cite{williams_aggressive_2016, williams_information_2017} as a model-predictive controller. We follow the standard formulation of MPPI, which requires us to specify 1) a dynamics model, 2) an action sampler, and 3) a cost function.

Our learned models serve as the dynamics model that MPPI optimizes over. Following recent advances in action-sampling techniques for MPPI \cite{bhardwaj2022storm, vlahov2024low}, we generate candidate action sequences via additive noise generated from a set of Gaussian walks with varying $\sigma$. We also add a fixed set of uniform action sequences to enable the robot to more effectively space potential local minima in the cost function. We grid sample these sequences uniformly in [throttle, steer] space, resulting in 441 uniform sequences that are evaluated at every MPPI iteration.

\begin{figure}[t]
    \centering
    \begin{minipage}{0.48\textwidth}
        \centering
        \begin{tabular}{c|c|c}
            \toprule
            Sampler & $\sigma$ & K \\
            \midrule
            Gaussian Walk 1 & [0.001, 0.001] & 128 \\
            Gaussian Walk 2 & [0.0005, 0.0005] & 128 \\
            Gaussian Walk 3 & [0.000025, 0.00025] & 128 \\
            Action Library & -- & 441 \\
            \bottomrule
        \end{tabular}
        \captionof{table}{Action Sampling Parameters}
        \label{tab:action_sample_params}
    \end{minipage}
    \hfill
    \begin{minipage}{0.48\textwidth}
        \centering
        \begin{tabular}{l|l}
            \toprule
            Parameter & Value \\
            \midrule
            $a_{\text{max}}$ ($m/s^2$) & $5.0$ \\
            $v_{\text{max}}$ ($m/s$) & $8.0$ or $10.0$ \\
            $K_\text{track}$ & 10.0 \\
            $K_\text{progress}$ & 1.0 \\
            \bottomrule
        \end{tabular}
        \captionof{table}{MPPI cost function parameters}
        \label{tab:mppi_cfn_params}
    \end{minipage}
\end{figure}

Our cost function for MPPI uses a combination of path-tracking costs and physics-derived safety limits (Eq. \ref{eq:mppi_cfn}). 
In a nutshell, this cost function encourages the robot to make forward progress along the reference path $\traj_\text{ref}$ while adhering to its shape (Eqs. \ref{eq:path_prog}, \ref{eq:path_track}), subject to speed and lateral acceleration constraints (Eqs. \ref{eq:speed_lim}, \ref{eq:lat_acc_lim}). Path progress can be interpreted as the amount of reference path (in $m$) traveled from the candidate trajectory's initial state $x_t$ to terminal state $x_{t+\horizon}$. Path track error can be interpreted as the average distance of the states in the trajectory to the reference path. The tradeoff between progress and accuracy is governed by the weights $K_\text{track}$ and $K_\text{progress}$. $K_\text{track}$ is set significantly higher than $K_\text{progress}$ to disincentivize cutting corners; however, as witnessed in \Cref{sec:online-results}, more tuning could be done to avoid MPPI-related failures and expose only dynamics modeling failures. Paths are represented as a dense sequence of $\text{SE}(2)$ poses and all path computation occurs in 2D (Eqs. \ref{eq:path_prog}, \ref{eq:path_track}). Hyperparameters are presented in Table \ref{tab:mppi_cfn_params}.

\begin{multline}
    \label{eq:mppi_cfn}
    J(\traj) = 
    J_{\text{speed}}(\traj, v_\text{max}) +
    J_{a_y}(\traj, a_\text{max}) +
    K_\text{track}J_\text{track}(\traj, \traj_\text{ref}) +
    K_\text{progress} J_\text{progress}(\traj, \traj_\text{ref})
\end{multline}

\begin{align}
    J_\text{speed}(\traj,v_\text{max}) &= \sum_{\state \in \traj} \left[
    \begin{cases}
    \infty \hbox{ if } \state_v > v_\text{max} \label{eq:speed_lim} \\
    0 \hbox{ otherwise} 
    \end{cases}
    \right] \\
    J_{a_y} &= \sum_{\state \in \traj} \left[
    \begin{cases}
    \infty \hbox{ if } \frac{\state_v^2 \tan(\delta)}{L} > a_\text{max} \label{eq:lat_acc_lim} \\
    0 \hbox{ otherwise} 
    \end{cases}
    \right] \\
    J_\text{progress}(\traj, \traj_{\text{ref}}) &= \sum^{i < i_{\text{end}}-1}_{i = i_{\text{start}}} ||\traj^{(i+1)}_{\text{ref}} - \traj^{(i)}_{\text{ref}}||_2 \label{eq:path_prog} \\
    & \quad i_{\text{start}} = \argmin_i \left[ ||\traj^{(i)}_{\text{ref}} - \state_t||_2 \right] \nonumber \\
    & \quad i_{\text{end}} = \argmin_i \left[ ||\traj^{(i)}_{\text{ref}} - \state_{t+\horizon}||_2 \right] \nonumber \\
    J_\text{track}(\traj, \traj_{\text{ref}}) &= \frac{1}{\horizon} \sum_{\state \in \traj} \left[ \min_{\state_{\text{ref}} \in \traj_{\text{ref}}} \left[ ||\state_{\text{ref}} - \state||_2 \right] \right] \label{eq:path_track}
\end{align}

\subsection{Hardware}
\label{sec:hardware}
\paragraph{Actuators}
The vehicle's action space covers steering and throttle, with no brake actuators available. It is important to note while autonomous action is relatively smooth, teleoperation actions are large discrete, because operators are controlling a gamepad controller with low-friction joysticks. 

Additionally, due to the resistance of off-road terrain and limitations of the steering DC motor, there is steering actuation delay, with $\sim$5s of delay from commanding max steer to reaching max steer. The dynamics model implicitly learns this steering delay. There are also standard throttle dynamics, as for any vehicle. 

\paragraph{Sensors}
For this work, the robot is equipped with a Velodyne VLP-32 LiDAR and a Multisense S21 RGB stereo camera. It is also has a MicroStrain 3DMGQ7-GNSS for GPS localization with RTK corrections. The LiDAR and camera run at 10Hz, whereas the RTK runs at 25Hz.

\paragraph{Basestation}
We deployed our perception and control algorithms on a laptop, equipped with an Intel Core Ultra 9 275HX CPU with 24 cores and 32GB of RAM and a laptop-grade Nvidia GeForce RTX 5080 with 16GB of VRAM.

\subsection{MPPI Failure Analysis}\label{sec:app-sys-mppi-fail}
\begin{figure}
    \centering
    \includegraphics[width=1.0\linewidth]{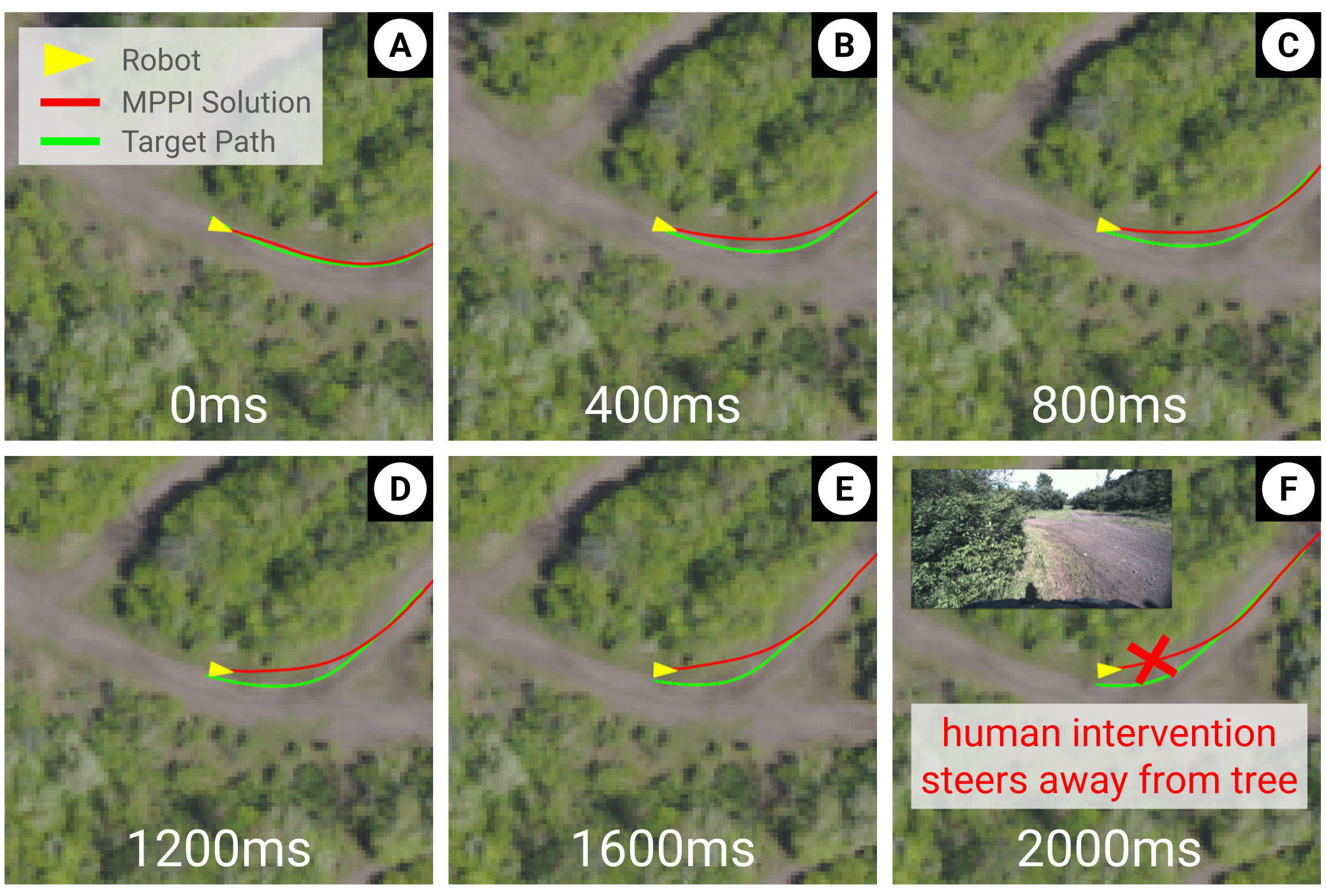}
    \caption{A robot following a target path with its optimal MPPI solution visualized. It is closely following the line in (A), highlighting accurate dynamics and control, but from (B) onwards, MPPI determines the optimal path undercuts the target path, nearly resulting in collision with dense vegetation.}
    \label{fig:mppi-failure}
\end{figure}

We detail why our model intervention from~\Cref{fig:online-qual} is not a modeling failure but instead a control failure due to the cost function tuning. Consider the scene in~\Cref{fig:mppi-failure} where each frame progresses after \unit{400}{ms} and the red MPPI solution is a new solution computed online. (A) shows the robot is following the target path at speed while closely tracking the reference line. Next, \unit{400}{ms} later in (B), while still closely following the line, MPPI determines its optimal solution undercuts the reference line, which begins the cause for our intervention. In (C-E), MPPI follows through with its undercut of the reference line, and our accurate dynamics yield little deviation. After \unit{2000}{ms} of following these undercutting solutions in (F), the human must intervene as the robot nears too dense of vegetation.

This particular intervention is most likely a cause of the tracking cost term weight, $K_\text{track}$, being too low and also due to sampling sparsity, where there was no action sequence that yielded a sample close enough to the reference path. This case study highlights how not all failures are dynamics modeling failures and how an MPC objective function can be very sensitive to its tunable parameters.

\section{Learned Dynamics Model Implementation}
\subsection{Network Architecture}
The base LSTM architecture is the same for all LSTM methods. The base architecture, composed of the context and predictor LSTM both have a hidden size $d_\rnnstate$ of 128. All methods use the same MLP state encoder and decoder, encoding to and from $d_\rnnstate$. All methods use the same MLP input encoders. For the predictor LSTM, the input encoder is purely an action encoder, encoding to $d_\rnninput^p$ of 16; however, for the context LSTM, the input encoder encodes state-action pairs, encoding to $d_\rnninput^c$ of 144. The context LSTM receives the encoded last state of the history buffer for hidden state initialization. Hyperparameters are in Table \ref{tab:hyperparams}.

\paragraph{Perception encoding}
The vanilla perception methods encode the perception once into a hidden state of dimension $d_\rnnstate$ to initialize the context LSTM. This latent gets concatenated with the state latent to form the initialization. The FPV LSTM uses a CNN encoder with $3\times$ downsampling blocks. The BEV LSTM uses a CNN encoder with $4\times$ downsampling blocks. These blocks apply the convolution, apply batch norm layers, use ReLU activation, and perform global max pooling.

Our spatial method queries the BEV map at the current predicted state. It queries 40 points, representing tire track points. This tensor of features $40\times F$ gets encoded with an MLP into the bev context latent size $d_\bevcontext$ of 32. This gets concatenated with the latent input to condition the next state prediction, and it gets concatenated with the hidden input and then added to apply the correction step. Hyperparameters are in~\Cref{tab:hyperparams}.

\begin{table}[]
    \centering
    \begin{tabular}{l|c|c}
        \toprule
        Hyperparameter & Variable & Value \\
        \midrule
        LSTM hidden size & $d_\rnnstate$ & 128 \\
        Predictor input encoding dim & $d_\rnninput^p$ & 16 \\
        Context input encoding dim & $d_\rnninput^c$ & 144 \\
        BEV context latent size & $d_\bevcontext$ & 32 \\
        Spatial query points (tire track) & $\mathcal{F}$ & 40 \\
        FPN CNN Downsample block & -- & $3\times$ \\
        \badbev CNN Downsample block & -- & $4\times$ \\
        BEV Features & $F$ & 36 \\
        Visual BEV Features & $F_{VFM}$& 32 \\
        Terrain BEV Features & $F_{Terrain}$ & 4 \\
        \bottomrule
    \end{tabular}
    \caption{Hyperparameters for the LSTM architecture and perception encoders.}
    \label{tab:hyperparams}
\end{table}

\subsection{BEV Footprint}\label{sec:app-net-footprint}

\begin{figure}
    \centering
     \begin{subfigure}[b]{0.4\textwidth}
         \centering
        \includegraphics[width=1.0\linewidth]{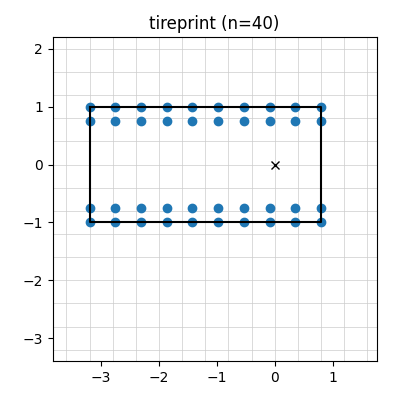}
        \caption{}
        \label{fig:app-footprints-nonrotated}
     \end{subfigure}
     \begin{subfigure}[b]{0.4\textwidth}
         \centering
        \includegraphics[width=1.0\linewidth]{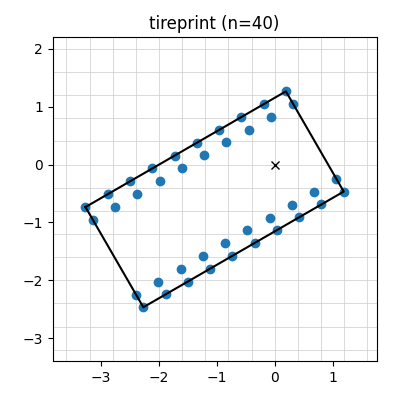}
        \caption{}
        \label{fig:app-footprints-rotated}
     \end{subfigure}
    \caption{Tire track footprint visualization over a BEV map of the \unit{100\times100\times0.4}{m} resolution aligned with the map (a) and rotated (b) to show how many new cells are queried. Vehicle outline is in black.}
    \label{fig:app-footprints}
\end{figure}

Our footprint is two ``tracks,'' separated \unit{1}{m} laterally from vehicle frame and uniformly spanning \unit{4\times0.42}{m} with \unit{10\times2}{points}. With a \unit{0.4}{m} BEV resolution, most points query unique cells, and we bi-linearly interpolate to prevent duplicate queries. \Cref{fig:app-footprints} shows our footprint over a BEV map aligned with the map and rotated. We do not condition on velocity or steer---our footprint aligns with our predicted position and orientation---nor suspension or roll, which follow from the terrain estimate. Slip and weight transfer are present in our training data and are represented implicitly in the recurrent latent, not as explicit state, since MPPI does not require them as planning states.

\subsubsection{Ablation Footprints}
\label{sec:app-net-footprint-ablation}

\begin{figure}
    \centering
    \includegraphics[width=0.8\linewidth]{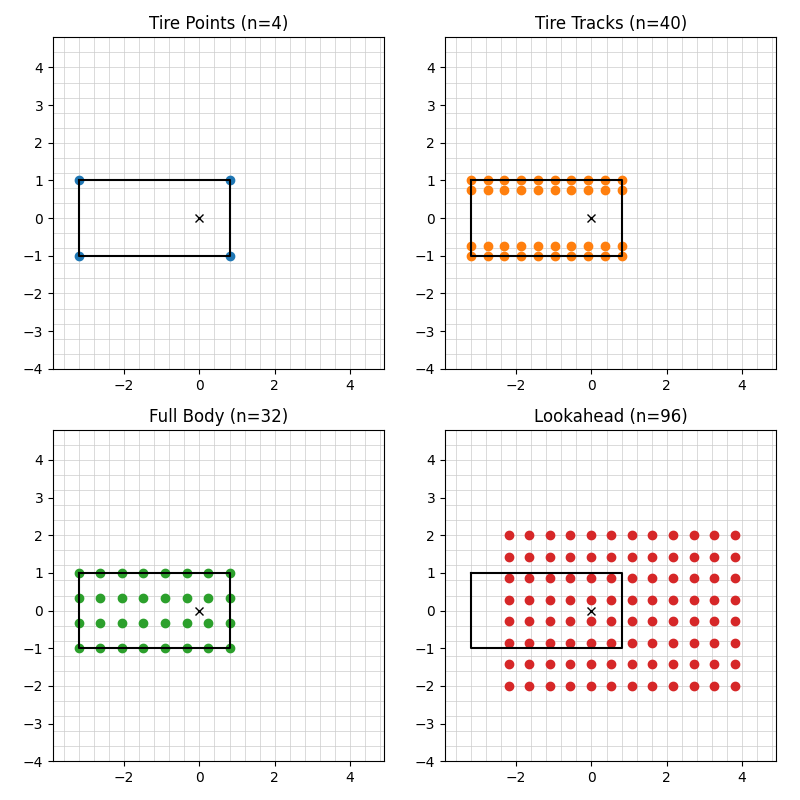}
    \caption{Footprints used for footprint ablation. Vehicle outline is in black.}
    \label{fig:app-footprint-ablation}
\end{figure}

For our ablation in~\Cref{sec:exp-offline}, we compare against these three other footprints, including our tracks footprint, shown in~\Cref{fig:app-footprint-ablation}. 4-point is the simplest footprint and has been used  by \citet{gibson_dynamics_2024}. Full Body is a naive footprint, without considering the tire location. And Lookahead is a footprint used by~\citet{lee_learning_2023} as a means to remove the prediction component of the perception querying by encompassing a larger area of where the robot can feasibly go.

\subsection{Activation Maps}
\label{sec:app-net-gradcam}
\begin{figure}[t]
    \centering
    \includegraphics[width=1.0\linewidth]{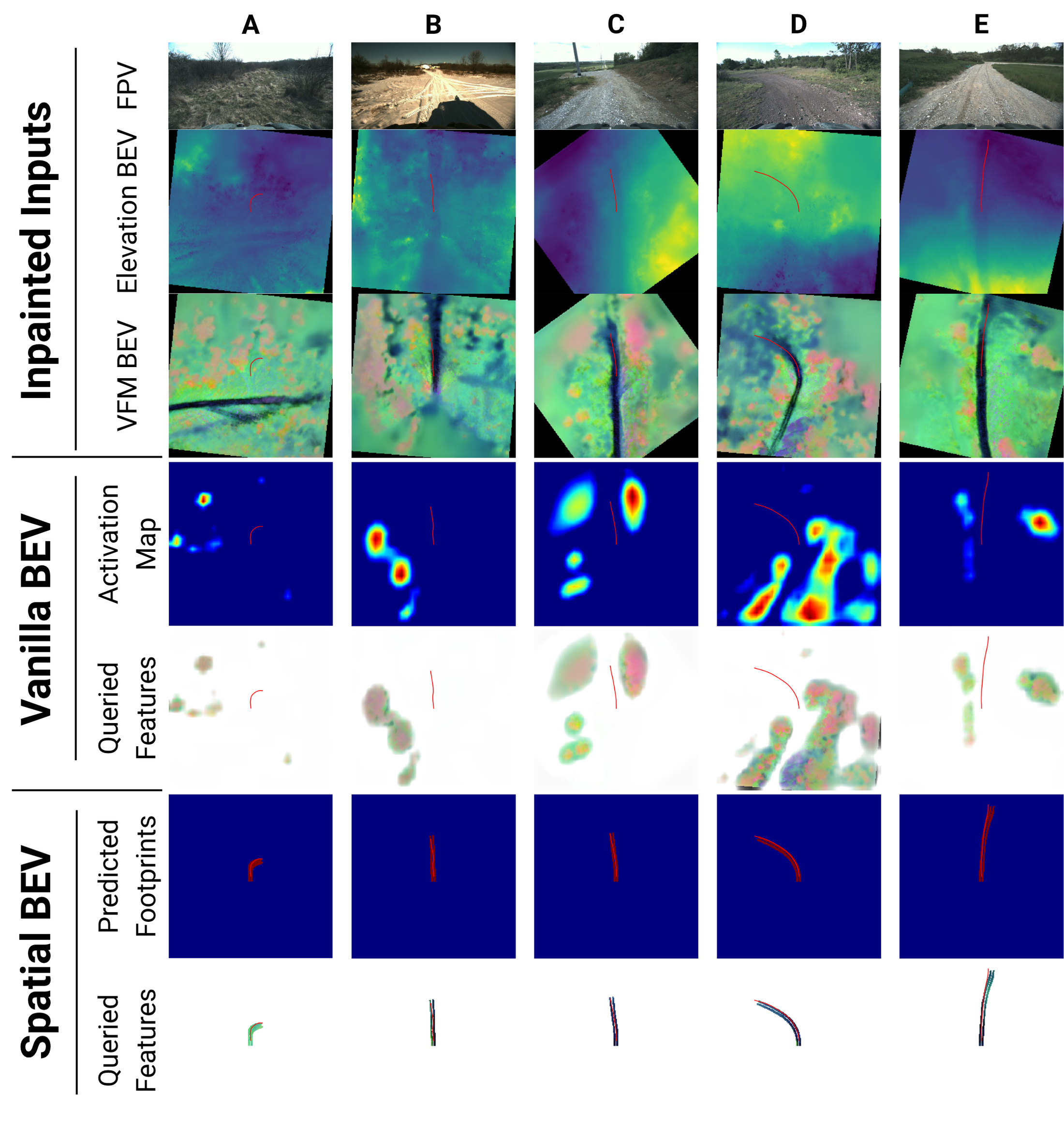}
    \caption{ Regions of interest for Vanilla BEV and our Spatial BEV approach on five different samples. Vanilla BEV attends to non-sensical regions for predicting motion, whereas our approach queries exactly where it predicts the robot to be, yielding a more informed environment context. The red line is the ground truth trajectory on all BEV plots. (A) through (E) show various environments and motions.}
    \label{fig:actmap-qual-more}
\end{figure}

We create these activation maps using GradCAM~\cite{selvaraju2017gradcam}, where we compute the activation needed to yield a single predicted scalar. In our case, for a given sample, we compute our prediction through the forward dynamics RNN. Then, we take the first predicted state's $x$ position, and back-propagate this result back to the last layer of our \badbev CNN encoder. At this layer, we compute the global importance of each channel by global-average-pooling its gradients, yielding a per-channel weight. We then compute a weighted vote by taking a per-cell weighted sum of the activations across channels, using these weights. To handle polarity, we apply a ReLU (or, alternately, an absolute value) to the weighted activation, retaining only features that positively (or, in the latter case, either positively or negatively) influence the predicted output. Finally, we upsample this coarse activation map to the input resolution to produce the final attention map seen in~\Cref{fig:actmap-qual}. We choose to back-propagate from the $x$ position because we saw, qualitatively, the activations were similar between state fields and along the rollout. More examples shown in~\Cref{fig:actmap-qual-more}.

\subsection{Training}
\paragraph{Training Data}
We train using state estimates from SuperOdometry \cite{zhao2021super}, a LiDAR-Inertial Odometry solution, but our models use RTK during real-world deployment. RTK was not available throughout the collection of this dataset, hence the mismatch. This does highlight performance robustness to state estimation solution.

\paragraph{Training Machines}
We trained our state-based methods on a single V100-32 GPU. Our BEV-based methods were trained on a single H100 GPU. While they could be trained on V100's, we used H100's for their superior VRAM, allowing us to have larger batches and more efficient training. It takes \unit{\mathord{\sim}30}{m} to train the state-based methods and \unit{\mathord{\sim}6}{hr} to train the BEV-based RNNs. The BEV-based transformer took about \unit{\mathord{\sim}24}{hr} to train.

\end{document}